\documentclass[twoside]{article}

\usepackage{hyperref}
\usepackage{amsmath,amssymb,amsthm,amsfonts,latexsym}        
\usepackage{url}
\usepackage{graphicx}
\usepackage{bm}
\usepackage{subfigure}
\usepackage{color}
\usepackage{appendix}
\usepackage{amsthm}
\usepackage{enumitem}

\usepackage[mathcal]{eucal}

\usepackage[accepted]{aistats2019}
\newcommand{\x}{w}
\newcommand{\xk}{w_{k}}
\newcommand{\mxk}{\bar{w}_{k}}
\newcommand{\xkk}{w_{k+1}}
\newcommand{\xopt}{w^{*}}

\newcommand{\vk}{v_{k}}
\newcommand{\yk}{\zeta_{k}}
\newcommand{\zk}{z_{k}}

\newcommand{\vkk}{v_{k+1}}

\newcommand{\rk}{r_{k}}
\newcommand{\rkk}{r_{k+1}}

\newcommand{\agk}{\alpha_{k}}
\newcommand{\bgk}{\beta_{k}}
\newcommand{\ak}{a_{k}}
\newcommand{\bk}{b_{k}}
\newcommand{\akk}{a_{k+1}}
\newcommand{\bkk}{b_{k+1}}
\newcommand{\gk}{\gamma_{k}}

\newcommand{\grad}[1]{\nabla f(#1)}

\newcommand{\sgrad}[2]{\nabla f(#1,#2)}

\newcommand{\norm}[1]{\left\|#1\right\|}
\newcommand{\normsq}[1]{\left\|#1\right\|^{2}}

\newcommand{\E}{\mathbb{E}}

\newcommand{\fj}{f_{i}}
\newcommand{\gradj}[1]{\nabla f_{i}(#1)}
\def \P{{\mathbb P}}

\newcommand{\transpose}{^\mathsf{\scriptscriptstyle T}}

\newtheorem{theorem}{Theorem}
\newtheorem{lemma}{Lemma}
\newtheorem{proposition}{Proposition}

\usepackage[backgroundcolor=white]{todonotes}

\begin{document}
\runningtitle{Fast and Faster Convergence of SGD for Over-Parameterized Models}
\runningauthor{Sharan Vaswani, Francis Bach, Mark Schmidt}

\twocolumn[
\aistatstitle{Fast and Faster Convergence of SGD for Over-Parameterized Models \\(and an Accelerated Perceptron)
}
\aistatsauthor{Sharan Vaswani$^{1}$ \And Francis Bach$^{2}$ \And Mark Schmidt$^{1}$}
\aistatsaddress{$^{1}$University of British Columbia \And $^{2}$INRIA, ENS, PSL Research University} 
]

\begin{abstract}
Modern machine learning focuses on highly expressive models that are able to fit or \emph{interpolate} the data completely,  resulting in zero training loss. For such models, we show that the stochastic gradients of common loss functions satisfy a \emph{strong growth condition}. Under this condition, we prove that constant step-size stochastic gradient descent (SGD) with Nesterov acceleration matches the convergence rate of the deterministic accelerated method for both convex and strongly-convex functions. 
We also show that this condition implies that SGD can find a first-order stationary point as efficiently as full gradient descent in non-convex settings. Under interpolation, we further show that all smooth loss functions with a finite-sum structure satisfy a \emph{weaker growth condition}. Given this weaker condition, we prove that SGD with a constant step-size attains the deterministic convergence rate in both the strongly-convex and convex settings. Under additional assumptions, the above results enable us to prove an $O(1/k^2)$ mistake bound for $k$ iterations of a stochastic perceptron algorithm using the squared-hinge loss. Finally, we validate our theoretical findings with experiments on synthetic and real datasets.  
\end{abstract}
\section{Introduction}
\label{sec:introduction}
Modern machine learning models are typically trained with iterative stochastic first-order methods~\cite{duchi2011adaptive,zeiler2012adadelta,kingma2014adam,schmidt2017minimizing,johnson2013accelerating,defazio2014saga}.
Stochastic gradient descent (SGD) and related methods such as Adagrad~\cite{duchi2011adaptive} or Adam~\cite{kingma2014adam} compute the gradient with respect to one or a mini-batch of training examples in each iteration and take a descent step using this gradient. Since these methods use only a small part of the data in each iteration, they are the preferred way for training models on large datasets. However, in order to converge to the solution, these methods require the step-size to decay to zero in terms of the number of iterations. This implies that the gradient descent procedure takes smaller steps as the training progresses. Consequently, these methods result in slow sub-linear rates of convergence. Specifically, if $k$ is the number of iterations, then SGD-like methods achieve a convergence rate of $O(1/k)$ and $O(1/\sqrt{k})$ for strongly-convex and convex functions respectively~\cite{nemirovski2009robust}. In practice, these methods are augmented with some form of momentum or acceleration~\cite{polyak1964some,nesterov2007gradient} that results in faster empirical convergence~\cite{sutskever2013importance}. Recently, there has been some theoretical analysis for the use of such acceleration in the stochastic setting~\cite{cohen2018acceleration}. Other related work includes algorithms specifically designed to achieve an accelerated rate of convergence in the stochastic setting~\cite{allen2017katyusha,lin2015universal,frostig2015regularizing}. 

Another recent trend in the literature has been to use variance-reduction techniques~\cite{schmidt2017minimizing,johnson2013accelerating,defazio2014saga} that exploit the finite-sum structure of the loss function in machine-learning applications. These methods do not require the step-size to decay to zero and are able to achieve the optimal rate of convergence. However, they require additional bookkeeping~\cite{schmidt2017minimizing,defazio2014saga} or need to compute the full gradient periodically~\cite{johnson2013accelerating}, both of which are difficult in the context of training complex models on large datasets. 

In this paper, we take further advantage of the optimization properties specific to modern machine learning models. In particular, we make use of the fact that models such as non-parametric regression or over-parameterized deep neural networks are expressive enough to fit or \emph{interpolate} the training dataset completely~\cite{zhang2016understanding,ma2018power}. For an SGD-like algorithm, this implies that the gradient with respect to each training example converges to zero at the optimal solution. This property of interpolation is also true for boosting~\cite{schapire1998boosting} and for simple linear classifiers on separable data. For example, the perceptron algorithm~\cite{rosenblatt1958perceptron} was first shown to converge to the optimal solution under a linear separability assumption on the data~\cite{novikoff1963convergence}. This assumption implies that the linear perceptron is able to fit the complete dataset without errors.

There has been some related work that takes advantage of the interpolation property in order to obtain faster rates of convergence for SGD~\cite{schmidt2013fast, ma2018power,cevher2018linear}. Specifically, Schmidt and Le Roux~\cite{schmidt2013fast} assume a \emph{strong growth condition} on the stochastic gradients. This condition relates the $\ell_{2}$ norms of the stochastic gradients to that of the full gradient. Under this assumption, they prove that constant step-size SGD can attain the same convergence rates as full gradient descent in both the strongly-convex and convex cases. Other related work has used the strong growth condition to prove convergence rates for incremental gradient methods~\cite{solodov1998incremental, tseng1998incremental}. Ma et al.~\cite{ma2018power} show that under weaker conditions, SGD with constant step-size results in linear convergence for strongly-convex functions. They also investigate the effect of batch-size on the convergence and theoretically justify the \emph{linear-scaling rule} used for training deep learning models in practice~\cite{goyal2017accurate}. Recently, Cevher and V\~{u} showed the linear convergence of proximal stochastic gradient descent under a weaker growth condition for restricted strongly convex functions~\cite{cevher2018linear}. They also analyse the effect of an additive error term on the convergence rate. 


In contrast to the above mentioned work, we first show that the strong growth condition (SGC)~\cite{schmidt2013fast} implies that SGD with a constant step-size and Nesterov momentum~\cite{nesterov2007gradient} achieves the accelerated convergence rate of the deterministic setting for both strongly-convex and convex functions (Section~\ref{sec:nesterov-acc}). Our result gives some theoretical justification behind the empirical success of using Nesterov acceleration with SGD~\cite{sutskever2013importance}. Further, in Section~\ref{sec:sgd-nc} we consider non-convex objectives and prove  under the SGC that constant step-size SGD  is able to find a first-order stationary point as efficiently as deterministic gradient descent. 
To the best of our knowledge, this is the first work to study accelerated and non-convex rates under the SGC. 

After the release of the first version of this work, Liu et al.~\cite{liu2018mass} also considered minimizing strongly-convex loss functions using a variant of Nesterov acceleration assuming interpolation. In this setting they show accelerated rates for the squared loss, and under additional assumptions give accelerated rates for general strongly-convex functions.  However, it is not clear if these additional assumptions are satisfied by common loss functions. Indeed, these additional assumptions imply the SGC (see Section~\ref{sec:wgc-practice}) so the result presented in Section~\ref{sec:nesterov-acc} is more widely-applicable.  Similarly, the work of Jain et al.~\cite{jain2017accelerating} uses tail-averaging to obtain accelerated rates but only for the special case of the squared loss under interpolation. Furthermore, unlike these works, we show accelerated rates for convex functions (that are not strongly-convex) under the SGC.

Another work appearing after the release of the initial version of this work is Bassily et al.~\cite{bassily2018exponential}, who considered minimizing non-convex functions satisfying the Polyak-Lojasiewicz~\cite{polyak1963gradient} (PL)  inequality (a generalization of strong-convexity) under the interpolation condition. This is a much stronger assumption than we make in Section~\ref{sec:sgd-nc} to analyze non-convex functions (since it implies all local optima are global optima), but under this condition they show that SGD can achieve a linear convergence rate. However, the step-size needed to achieve this rate is proportional to the PL constant which is typically extremely small (and is often is both unknown and difficult to estimate). By exploiting the stronger SGC, in this version of the paper we have added a result under the PL inequality (Section~\ref{sec:sgd-nc}) that achieves a faster rate by using a step-size that depends only on the smoothness properties of the functions.

In this work, we also relax the strong growth condition to a more practical \emph{weak growth condition} (WGC). In Section~\ref{sec:wgc}, we prove that the weak growth condition is sufficient to obtain the optimal convergence of constant step-size SGD for smooth strongly-convex and convex functions. To demonstrate the applicability of our growth conditions in practice, we first show that for models interpolating the data, the WGC is satisfied for all smooth and convex loss functions with a finite-sum structure (Section~\ref{sec:wgc-practice}). Furthermore, we prove that functions satisfying the WGC and the PL condition also satisfy the SGC. Under additional assumptions, we show that it is also satisfied for the squared-hinge loss. This result enables us to prove an $O(1/k^2)$ mistake bound for $k$ iterations of an accelerated stochastic perceptron algorithm using the squared-hinge loss (Section~\ref{sec:perceptron}). Finally, in Section~\ref{sec:experiments}, we evaluate our claims with experiments on synthetic and real datasets.

\section{Background}
\label{sec:background}
In this section, we give the required background and set up the necessary notation. Our aim is to minimize a differentiable function $f(\x)$. Depending on the context, this function can be strongly-convex, convex or non-convex. We assume that we have access to noisy gradients for the function $f$ and use stochastic gradient descent (SGD) for $k$ iterations in order to minimize it. The SGD update rule in iteration $k$ can be written as: $\xkk = \xk - \eta_{k} \sgrad{\xk}{\zk}$. Here, $\xkk$ and $\xk$ are the SGD iterates, $\zk$ is the gradient noise and $\eta_{k}$ is the step-size at iteration $k$. We assume that the gradients $\sgrad{\x}{z}$ are unbiased, implying that for all $\x$ and $z$ that $\E_{z} \left[ \sgrad{\x}{z} \right] = \grad{\x}$. 

While most of our results apply for general SGD methods, a subset of our results rely on the function $f(\x)$ having a finite-sum structure meaning that $f(\x) = \frac{1}{n} \sum_{i = 1}^{n} \fj(\x)$. In the context of supervised machine learning, given a training dataset of $n$ points, the term $\fj(w)$ corresponds to the loss function for the point $(x_i, y_i)$ when the model parameters are equal to $w$. Here $x_i$ and $y_i$ refer to the feature vector and label for point $i$ respectively. Common choices of the loss function include the squared loss where $\fj(\x) = \frac{1}{2}\left( \x \transpose x_i - y_i \right)^2$, the hinge loss where $\fj(\x) = \max(0, 1 - y_i \x \transpose x_i )$ or the squared-hinge loss where $\fj(\x) = \max \left( 0, 1 - y_i \x \transpose x_i \right)^{2}$. The finite sum setting includes both simple models such as logistic regression or least squares and more complex models like non-parametric regression and deep neural networks. 

In the finite-sum setting, SGD consists of choosing a point and its corresponding loss function (typically uniformly) at random and evaluating the gradient with respect to that function. It then performs a gradient descent step: $\xkk = \xk - \eta_{k} \nabla f_{k}(\xk)$ where $f_{k}(\cdot)$ is the random loss function selected at iteration $k$. The unbiasedness property is automatically satisfied in this case, i.e. $\E_i \left[ \gradj{\x} \right] = \grad{\x}$ for all $\x$. Note that in this case, the random selection of points for computing the gradient is the source of the noise $z_k$. In order to converge to the optimum, SGD requires the step-size $\eta_{k}$ to decrease with $k$; specifically at a rate of $\frac{1}{\sqrt{k}}$ for convex functions and at a $\frac{1}{k}$ rate for strongly-convex functions. Decreasing the step-size with $k$ results in sub-linear rates of convergence for SGD. 

In order to derive convergence rates, we need to make additional assumptions about the function $f$~\cite{nemirovski2009robust}. Beyond differentiability, our results assume that the function $f(\cdot)$ satisfies some or all of the following common assumptions. For all points $w$, $v$ and for constants $f^*$, $\mu$, and $L$; 
\begin{align*}
f(w) &\geq f^* & \tag{Bounded below}\\
f(v) & \geq f(w) + \langle \grad{w}, v - w \rangle & \tag{Convexity} \\
f(v) & \geq f(w) + \langle \grad{w}, v - w \rangle + \frac{\mu}{2} \normsq{v - w} & \tag{$\mu$ Strong-convexity} \\
f(v) & \leq f(w) + \langle \grad{w}, v - w \rangle + \frac{L}{2} \normsq{v - w} & \tag{$L$ Smoothness} 
\end{align*}
Note that some of our results in Section~\ref{sec:practice} rely on the finite-sum structure and we explicitly state when we need this additional assumption. 

In this paper, we consider the case where the model is able to \emph{interpolate} or fit the labelled training data completely. This is true for expressive models such as non-parametric regression and over-parametrized deep neural networks. For common loss functions that are lower-bounded by zero, interpolating the data results in zero training loss. Interpolation also implies that the gradient with respect to each point converges to zero at the optimum. Formally, in the finite-sum setting, if the function $f(\cdot)$ is minimized at $\xopt$, i.e., if $\grad{\xopt} = 0$, then for all functions $\fj(\cdot)$, $\gradj{\xopt} = 0$. 

The \emph{strong growth condition} (SGC) used  connects the rates at which the stochastic gradients shrink relative to the full gradient. Formally, for any point $\x$ and the noise random variable $z$, the function $f$ satisfies the strong growth condition with constant $\rho$ if, 
\begin{align}
\E_{z} \normsq{\sgrad{\x}{z}} & \leq \rho \normsq{\grad{\x}}. \label{eq:sgc-main} \\
\intertext{Equivalently, in the finite-sum setting,}
\E_{i} \normsq{\gradj{\x}} & \leq \rho \normsq{\grad{\x}}. \label{eq:sgc-main-finite} 
\end{align}
For this inequality to hold, if $\grad{\x} = 0$, then $\gradj{\x} = 0$ for all $i$. Thus, functions satisfying the SGC necessarily satisfy the above interpolation property. Schmidt and Le Roux's work~\cite{schmidt2013fast} derives optimal convergence rates for constant step-size SGD under the above condition for both convex and strongly-convex functions. In the next section, we show that the SGC implies the accelerated rate of convergence for constant step-size SGD with Nesterov momentum. 
\section{SGD with Nesterov acceleration under the SGC}
\label{sec:nesterov-acc}
We first describe constant step-size SGD with Nesterov acceleration. The algorithm consists of three sequences ($\xk,\yk,\vk$) updated in each iteration~\cite{nesterov2012efficiency}. Specifically, it consists of the following update rules:
\begin{align}
\xkk &= \yk - \eta \sgrad{\yk}{\zk} \label{eq:grad-step} \\
\yk  &= \agk \vk + (1 - \agk)\xk \label{eq:mix-step} \\
\vkk &= \bgk \vk + (1 - \bgk) \yk - \gk \eta \sgrad{\yk}{\zk}. \label{eq:egrad-step} 
\end{align}
Here, $\eta$ is the constant step-size for the SGD step and $\agk$, $\bgk$, $\gk$ are tunable parameters to be set according to the properties of $f$. 

In order to derive a convergence rate for the above algorithm under the SGC, we first observe that a form of the SGC is satisfied in the case of coordinate descent~\cite{wright2015coordinate}. In this case, we choose a coordinate (typically at random) and perform a gradient descent step with respect to that coordinate. The notion of a coordinate in this case is analogous to that of an individual loss function in the finite sum case. For coordinate descent, a zero gradient at the optimal solution implies that the partial derivative with respect to each coordinate is also equal to zero. This is analogous to the SGC in the finite-sum case, although we note the results in this section do not require the finite-sum assumption.

We use this analogy formally in order to extend the proof of Nesterov's accelerated coordinate descent~\cite{nesterov2012efficiency} to derive convergence rates for the above algorithm when using the SGC. This enables us to prove the following theorems (with proofs in Appendices~\ref{app:sc-a} and~\ref{app:c-a}) in both the strongly-convex and convex settings. 

\begin{theorem}[Strongly convex]
Under $L$-smoothness and $\mu$ strong-convexity, if $f$ satisfies the SGC with constant $\rho$, then SGD with Nesterov acceleration with the following choice of parameters,  
\begin{flalign*}
\gk  & = \frac{1}{\sqrt{\mu \eta \rho}} \quad \text{;} \quad \bgk = 1 - \sqrt{\frac{\mu \eta}{\rho}} \\
\bkk & =  \frac{\sqrt{\mu}}{\left( 1 - \sqrt{\frac{\mu \eta}{\rho}} \right)^{(k+1)/2}} \\
\akk & = \frac{1}{\left( 1 - \sqrt{\frac{\mu \eta}{\rho}} \right)^{(k+1)/2}} \\ 
\agk & = \frac{\gk \bgk \bkk^2 \eta}{\gk \bgk \bkk^2 \eta + \ak^2} \text{;} \quad \eta = \frac{1}{\rho L} ,
\end{flalign*}
results in the following convergence rate:
\begin{flalign*}
& \E f(\xkk) - f(\xopt) \\
& \leq \left( 1 - \sqrt{\frac{\mu}{\rho^2 L}} \right)^{k} \left[ f(\x_0) - f(\xopt) + \frac{\mu}{2} \normsq{\x_0 - \xopt} \right] .
\end{flalign*}
\label{thm:sc-a}
\end{theorem}

\begin{theorem}[Convex]
Under $L$-smoothness and convexity, if $f$ satisfies the  SGC with constant $\rho$, then SGD with Nesterov acceleration with the following choice of parameters,  
\begin{flalign*}
\gk  & = \frac{\frac{1}{\rho} + \sqrt{\frac{1}{\rho^2} + 4 \gamma_{k-1}^2}}{2} \\
\akk & = \gk \sqrt{\eta \rho} \\
\agk & = \frac{\gk \eta}{\gk \eta + \ak^2} \text{;} \quad \eta = \frac{1}{\rho L},
\end{flalign*}
results in the following convergence rate:
\begin{flalign*}
\E f(\xkk) - f(\xopt) & \leq \frac{2 \rho^2 L}{k^2} \normsq{\x_0 - \xopt}.
\end{flalign*}
\label{thm:c-a}
\end{theorem}
The above theorems show that constant step-size SGD with Nesterov momentum achieves the accelerated rate of convergence up to a $\rho^2$ factor for both strongly-convex and convex functions.  

In Appendix~\ref{app:additive}, we consider the SGC with an extra additive error term, resulting in the following condition: $\E_{z} \normsq{\sgrad{\x}{z}} \leq \rho \normsq{\grad{\x}} + \sigma^2$. We analyse the rate of convergence of the above algorithm under this modified condition and obtain a similar dependence on $\sigma$ as in Cohen et al.~\cite{cohen2018acceleration}. 
\section{SGD for non-convex functions satisfying the SGC}
\label{sec:sgd-nc}
In this section, we show that the SGC results in an improvement over the $O\left(1/\sqrt{k}\right)$ rate for SGD in the non-convex setting~\cite{ghadimi2013stochastic}. In particular, we show that under the strong growth condition, constant step-size SGD is able to find a first-order stationary point as efficiently as deterministic gradient descent. We prove the following theorem (with the proof in Appendix~\ref{app:nc-na}),
\begin{theorem}[Non-Convex]
Under $L$-smoothness, if $f$ satisfies SGC with constant $\rho$, then SGD with a constant step-size $\eta = \frac{1}{\rho L}$ attains the following convergence rate:
\begin{align*}
\min_{i = 0,1, \ldots k-1} \E \left[ \normsq{\grad{w_i}} \right] \leq \left( \frac{2 \rho L}{k} \right) \left[ f(\x_{0}) - f^* \right].
\end{align*}
\label{thm:nc-na}
\end{theorem}
The above theorem shows that under the SGC, SGD with a constant step-size can attain the optimal $O(1/k)$ rate for non-convex functions. To the best of our knowledge, this is the first result for non-convex functions under interpolation-like conditions. Under these conditions, constant step-size SGD has a better convergence rate than algorithms which have recently been proposed to improve on SGD~\cite{allen2017natasha,carmon2017convex}. Note that the above theorem applies to neural networks with a sigmoid activation function under the assumption that the strong-growth condition is satisfied. Hence, our results also provide some theoretical justification for the effectiveness of SGD for non-convex over-parameterized models like deep neural networks. 

Under the additional assumption that the function satisfies the Polyak- Lojasiewicz condition~\cite{polyak1963gradient} (a generalization of strong-convexity), we show that SGD can obtain linear convergence. Specifically, we prove the following theorem (with the proof in Appendix~\ref{app:nc-pl-na}),
\begin{theorem}[Non-Convex + PL]
Under $L$-smoothness, if $f$ satisfies SGC with constant $\rho$ and the Polyak- Lojasiewicz inequality with constant $\mu$, then SGD with a constant step-size $\eta = \frac{1}{\rho L}$ attains the following convergence rate:
\begin{align*}
\E \left[ f(\xkk) - f^* \right] & \leq \left(1 - \frac{\mu}{\rho L} \right)^{k} \left[ f(\x_{0}) - f^* \right].
\end{align*}
\label{thm:nc-pl-na}
\end{theorem}
Note that the PL condition or a related notion of restricted strong-convexity (RSI)~\cite{karimi2016linear} is satisfied by numerous non-convex optimization problems of interest. These include neural networks~\cite{li2017convergence, kleinberg2018alternative, soltanolkotabi2018theoretical}, matrix completion~\cite{sun2016guaranteed} and phase retrieval~\cite{chen2015solving}. Under the additional SGC assumption, the above theorem implies fast rates of convergence for SGD on these problems. In contrast, Bassily et al.~\cite{bassily2018exponential} do not assume the SGC and achieve a rate of $\left(1 - \frac{\mu^2}{L^2} \right)^{k}$ using a much smaller step-size $\eta = \frac{\mu}{L^2}$. 

\section{Weak growth condition}
\label{sec:wgc}
In this section, we relax the strong growth condition to a more practical condition which we refer to as the \emph{weak growth condition} (WGC). Formally, if the function $f(\cdot)$ is $L$-smooth and has a minima at $\xopt$, then it satisfies the WGC with constant $\rho$, if for all points $\x$ and noise random variable~$z$,
\begin{align}
\E_{z} \normsq{\sgrad{\x}{z}} & \leq 2 \rho L [ f(\x) - f(\xopt) ] .\label{eq:wgc-main} 
\intertext{Equivalently, in the finite-sum setting,}
\E_{i} \normsq{\gradj{\x}} & \leq 2 \rho L [ f(\x) - f(\xopt) ] .\label{eq:wgc-main-finite} 
\end{align}
In the above condition, notice that if $\x = \xopt$, then $\gradj{\xopt} = 0$ for all points $i$. Thus, the WGC implies the interpolation property explained in Section~\ref{sec:background}. 

\subsection{Relation between WGC and SGC}
\label{sec:sgc-wgc}
In this section, we relate the two growth conditions. We first prove that SGC implies WGC with the same $\rho$ without any additional assumptions, formally showing that the WGC is indeed weaker than the corresponding SGC. For the converse, a function satisfying the WGC satisfies the SGC with a worse constant if it also satisfies the Polyak- Lojasiewicz (PL) inequality~\cite{polyak1963gradient}. The above relations are captured by the following proposition, proved in Appendix~\ref{app:wgc-sgc}
\begin{proposition}
If $f(\cdot)$ is $L$-smooth, satisfies the WGC with constant $\rho$ and the PL inequality with constant $\mu$, then it satisfies the SGC with constant $\frac{\rho L}{\mu}$. 

Conversely, if $f(\cdot)$ is $L$-smooth, convex and satisfies the SGC with constant $\rho$, then it also satisfies the WGC with the same constant $\rho$. 
\label{prop:wgc-sgc}
\end{proposition}

\subsection{SGD under the weak growth condition}
Using the WGC, we obtain the following convergence rates for SGD with a constant step-size. 
\begin{theorem}[Strongly-convex]
Under $L$-smoothness and $\mu$ strong-convexity, if $f$ satisfies the WGC with constant $\rho$, then SGD with a constant step-size $\eta = \frac{1}{\rho L}$ achieves the following rate:
\begin{align*}
\E \normsq{\xkk - \xopt} & \leq \left(1 - \frac{\mu}{\rho L} \right)^{k} \normsq{\x_0 - \xopt}.
\end{align*}
\label{thm:sc-na}
\end{theorem}

\begin{theorem}[Convex]
Under $L$-smoothness and convexity, if $f$ satisfies the WGC with constant $\rho$, then SGD with a constant step-size $\eta = \frac{1}{4 \rho L}$ and iterate averaging achieves the following rate:
\begin{align*}
\E[f(\mxk)]  - f(\xopt) & \leq \frac{ 4 L \left(1 +\rho \right) \normsq{\x_0 - \xopt} }{k}.
\end{align*}
Here, $\mxk = \frac{\left[ \sum_{i = 1}^{k} \x_{i} \right]}{k}$ is the averaged iterate after $k$ iterations. 
\label{thm:c-na}
\end{theorem}
The proofs for Theorems~\ref{thm:sc-na} and~\ref{thm:c-na} are deferred to Appendices~\ref{app:sc-na} and~\ref{app:c-na} respectively. In these cases, the WGC is sufficient to show that constant step-size SGD can attain the deterministic rates up to a factor of $\rho$. Since this condition is weaker than the corresponding strong growth condition, our results subsume the SGC results~\cite{schmidt2013fast}. Note that an alternative way to obtain the result in Theorem~\ref{thm:sc-na} would be to observe that the WGC and strong convexity imply the SGC (with a constant $\frac{\rho L}{\mu}$) (Proposition 1) and then use the result by Schmidt et al.~\cite{schmidt2013fast}. This would result in an additional dependence on $\frac{\mu}{\rho L}$ which is worse than the rate in Theorem~\ref{thm:sc-na}. 

In the next section, we characterize the functions satisfying the growth conditions in practice. 
\section{Growth conditions in practice}
\label{sec:practice}
In this section, we give examples of functions that satisfy the weak and strong growth conditions. In Section~\ref{sec:wgc-practice}, we first show that for models interpolating the data, the WGC is satisfied by all smooth functions with a finite-sum structure. In section~\ref{sec:sgc-practice}, we show that the SGC is satisfied by the squared-hinge loss under additional assumptions. 

\subsection{Functions satisfying WGC}
\label{sec:wgc-practice}
To characterize the functions satisfying the WGC, we first prove the following proposition (with the proof in Appendix~\ref{app:wgc-f}):
\begin{proposition}
If the function $f(\cdot)$ is convex and has a finite-sum structure for a model that interpolates the data and $L_{\max}$ is the maximum smoothness constant amongst the functions $\fj(\cdot)$, then for all $\x$, 
\begin{align}
\E_i \normsq{\gradj{\x}} & \leq 2 L_{max} \left[ f(\x) - f(\xopt)  \right]. & \label{eq:belkin-wgc}
\end{align}
\label{prop:wgc-f}
\end{proposition}
Comparing the above equation to Equation~\ref{eq:wgc-main-finite}, we see that any smooth finite-sum problem under interpolation satisfies the WGC with $\rho = \frac{L_{max}}{L}$. The WGC is thus satisfied by common loss functions such as the squared and squared-hinge loss. For these loss functions, if $L_i = L$ for all $i$, then Theorem~\ref{thm:sc-na} implies that SGD with $\eta = \frac{1}{L}$ results in linear convergence for strongly-convex functions. This matches the recently proved result of Ma et al.~\cite{ma2018power}, whereas Theorem~\ref{thm:c-na} allows us to generalize their result beyond strongly-convex functions.  

\subsection{Functions satisfying SGC}
\label{sec:sgc-practice}
We now show that under additional assumptions on the data, the squared-hinge loss also satisfies the SGC. We first assume that the data is linearly separable with a margin equal to $\tau$, implying that for all $x$, $\tau = \max_{\vert \x \vert = 1} \inf_{x \in \mathcal{S}} \x^\top x$. Here, $\mathcal{S}$ is the support of the distribution of the features $x$. Note that the above assumption implies the existence of a classifier $\xopt$ such that $\vert \vert \xopt \vert \vert = \frac{1}{\tau}$. In addition to this, we assume that the features have a finite support, meaning that the set $\mathcal{S}$ is finite and has a 
cardinality equal to $c$. Under these assumptions, we prove the following lemma in Appendix~\ref{app:sq-hinge-sgc}, 
\begin{lemma}
For linearly separable data with margin $\tau$ and a finite support of size $c$, the squared-hinge loss satisfies the SGC with the constant $\rho = \frac{c}{\tau^2}$. 
\label{lemma:sq-hinge-sgc}
\end{lemma}
In the next section, we use the above lemma to prove a mistake bound for the perceptron algorithm using the squared-hinge loss. 
\section{Implication for Faster Perceptron}
\label{sec:perceptron}
In this section, we use the strong growth property of the squared-hinge function in order to prove a bound on the number of mistakes made by the perceptron algorithm~\cite{rosenblatt1958perceptron} using a squared-hinge loss. The perceptron algorithm is used for training a linear classifier for binary classification and is guaranteed to converge for linearly separable data~\cite{novikoff1963convergence}. It can be considered as stochastic gradient descent on the loss $\fj(\x) = \max \{0, y_i x_i^\top \x \}$. 

The common way to characterize the performance of a perceptron is by bounding the number of mistakes (in the binary classification setting) after $k$ iterations of the algorithm. In other words, we care about the quantity $\P ( y x^\top\xk \geqslant 0)$. Assuming linear separability of the data and that $\vert \vert x \vert \vert = 1$ for all points $(x,y)$, the perceptron achieves a mistake bound of $O\left(\frac{1}{\tau^2}\right)$~\cite{novikoff1963convergence}. 

In this paper, we consider a modified perceptron algorithm using the squared-hinge function as the loss. Note that since we assume the data to be linearly separable, a linear classifier is able to fit all the training data. Since the squared-hinge loss function is smooth, the conditions of Proposition~\ref{prop:wgc-f} are satisfied, which implies that it satisfies the WGC with $\rho = \frac{L_{max}}{L}$. Also observe that since we assume that $\vert \vert x \vert \vert = 1$, $L_{max} = L = 1$. Using these facts with Theorem~\ref{thm:c-na} and assuming that we start the optimization with $\x_0 = \mathbf{0}$, we obtain the following convergence rate using SGD with $\eta = 1/4$, 
\begin{align*}
\E[f(\xkk)] & \leq \frac{8}{\tau^2 k}.
\end{align*}
To see this, recall that $\vert \vert \xopt \vert \vert = \frac{1}{\tau}$ and the loss is equal to zero at the optima, implying that $f(\xopt) = 0$. 

The above result gives us a bound on the training loss. We use the following lemma (proved using the Markov inequality in Appendix~\ref{app:surrogate-01}) to relate the mistake bound to the training loss. 
\begin{lemma}
If $f(\x,x,y)$ represents the loss on the point $(x,y)$, then
\begin{align*}
\P ( y x^\top \x \leqslant 0) \leqslant \E_{x,y} f(\x,x,y).
\end{align*}
\label{lemma:surrogate-01}
\end{lemma}
Combining the above results, we obtain a mistake bound of $O\left(\frac{1}{\tau^2 k}\right)$ when using the squared-hinge loss on linearly separable data.  We thus recover the standard results for the stochastic perceptron. 

Note that for a finite amount of data (when the expectation is with respect to a discrete distribution), if we use batch accelerated gradient descent (which is not one of the stochastic gradient algorithms studied in this paper, and for which no growth condition is needed), we obtain a mistake bound that decreases as $1/k^2$. This improves on existing mistake bounds that scale as $1/k$~\cite{soheili2013primal,yu2014saddle}. Note that both sets of algorithms have the same dependence on the margin $\tau$, but this deterministic accelerated method would require evaluating $n$ gradients on each iteration.

From Lemma~\ref{app:surrogate-01}, we know that the squared-hinge loss satisfies the SGC with $\rho = \frac{c}{\tau^2}$. Under the same conditions as above, this lemma along with the result of Theorem~\ref{thm:c-a} gives us the following bound:
\begin{align*}
\E f(\xkk) & \leq \frac{2 c^2}{\tau^6 k^2} .
\end{align*}
Using the result from Lemma~\ref{lemma:surrogate-01}, this results in a mistake bound of the order $O\left(\frac{1}{\tau^6 k^2}\right)$ while only requiring one gradient per iteration. Hence, the use of acceleration leads to an improved novel dependence of $O(1/k^2)$, but requires the additional assumptions of Lemma~\ref{app:surrogate-01} and has a worse dependence on the margin $\tau$.
\section{Experiments}
\label{sec:experiments}
\begin{figure*}[!ht]
\centering
        \subfigure[$\tau = 0.1$]
        {
			\includegraphics[width=0.4\textwidth]{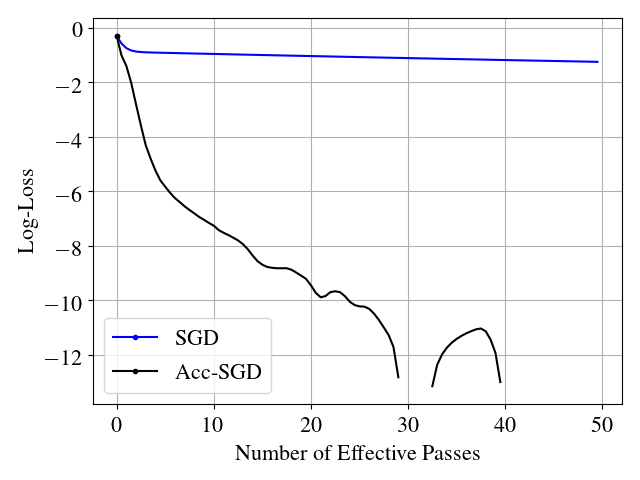}
			\label{fig:gamma-01}
        }        
        \subfigure[$\tau = 0.05$]
        {
			\includegraphics[width=0.4\textwidth]{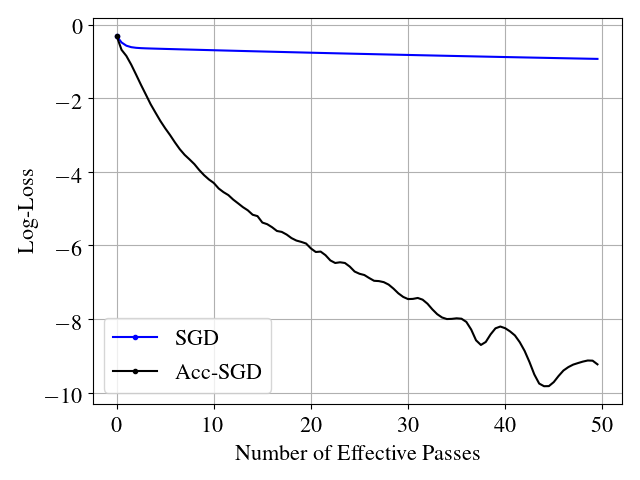}
			\label{fig:gamma-005}
        }      
        \\  
        \subfigure[$\tau = 0.01$]
        {
			\includegraphics[width=0.4\textwidth]{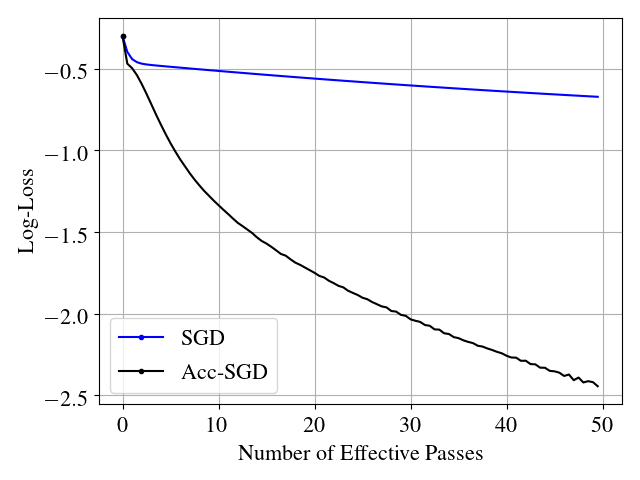}
			\label{fig:gamma-001}
        }        
        \subfigure[$\tau = 0.005$]
        {
			\includegraphics[width=0.4\textwidth]{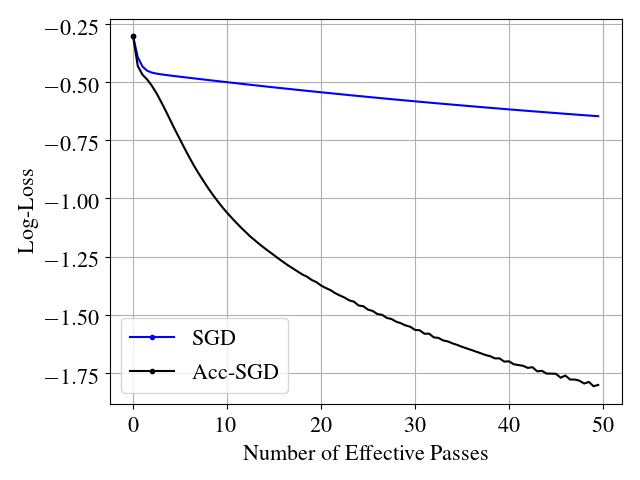}
			\label{fig:gamma-0005}
        }        
\caption{Comparison of SGD and variants of accelerated SGD on a synthetic linearly separable dataset with margin $\tau$. Accelerated SGD with $\eta = \tau/L$ leads to faster convergence as compared to SGD with $\eta = 1/L$. }
\label{fig:synthetic}
\end{figure*}    
\begin{figure*}[!ht]
\centering
        \subfigure[CovType]
        {
			\includegraphics[width=0.4\textwidth]{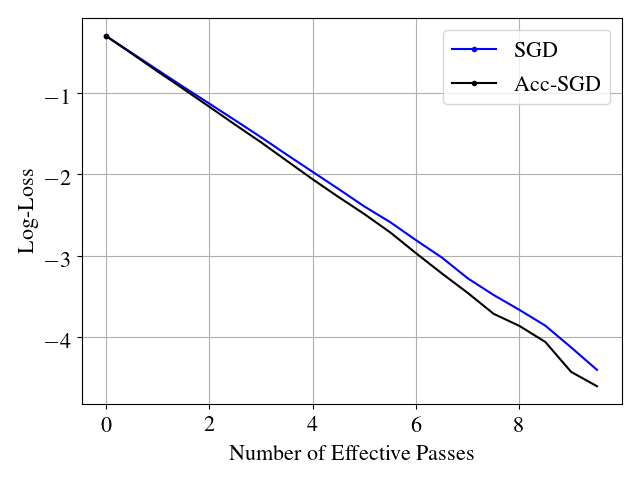}
			\label{fig:covtype}
        }        
        \subfigure[Protein]
        {
			\includegraphics[width=0.4\textwidth]{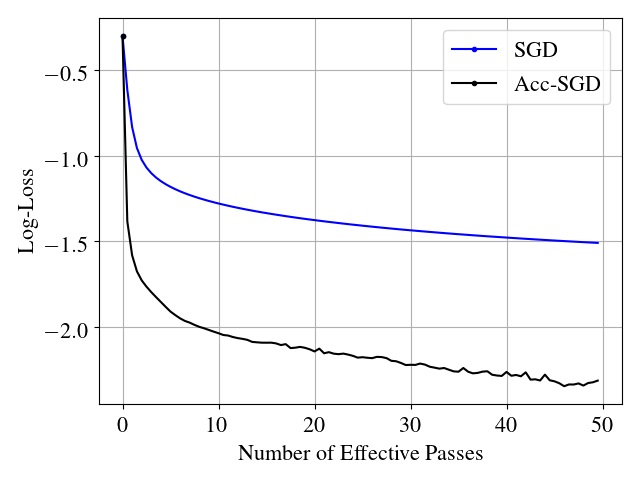}
			\label{fig:protein}
        }               
\caption{Comparison of SGD and accelerated SGD for learning a linear classifier with RBF features on the (a) CovType and (b) Protein datasets. Accelerated SGD leads to better performance as compared to SGD with $\eta = 1/L$.}
\label{fig:real}
\end{figure*}
In this section, we empirically validate our theoretical results. For the first set of experiments (Figures~\ref{fig:gamma-01}-\ref{fig:gamma-0005}), we generate a synthetic binary classification dataset with $n = 8000$ and the dimension $d = 100$. We ensure that the data is linearly separable with a margin $\tau$, thus satisfying the interpolation property for training a linear classifier. We seek to minimize the finite-sum squared-hinge loss, $f(\x) = \sum_{i = 1}^{n} \max \left( 0, 1 - y_i x_i\transpose \x \right)^{2}$. In Figure~\ref{fig:synthetic}, we vary the margin $\tau$ and plot the logarithm of the loss with the number of effective passes (one pass is equal to $n$ iterations of SGD) over the data. In all of our experiments, we estimate the value of the smoothness parameter $L$ as the maximum eigenvalue of the Gram matrix $X^{T}X$. 

We evaluate the performance of constant step-size SGD with and without acceleration. Since the squared-hinge loss satisfies the WGC with $\rho = \frac{L_{max}}{L}$ (Proposition~\ref{prop:wgc-f}), we use SGD with a constant step-size $\eta = 1 / L_{max}$\footnote{Note that using $\eta = 1/L_{max}$ lead to consistently better results as compared to using $\eta = 1/4 L_{max}$ as suggested by Theorem~\ref{thm:c-na}.} (denoted as SGD in the plots). For using Nesterov acceleration, we experimented with the dependence of the margin $\tau$ on the constant $\rho$ in the SGC. We found that setting $\rho = 1 /\tau$ results in consistently stable but fast convergence across different choices of $\tau$. We thus use a step-size $\eta = \tau / L$ and set the tunable parameters in the update Equations~\ref{eq:grad-step}-\ref{eq:egrad-step} as specified by Theorem~\ref{thm:c-a}. We denote this variant of accelerated SGD as Acc-SGD in the subsequent plots. In Appendix~\ref{app:exps}, we propose a line-search heuristic to dynamically estimate the value of $\rho$. 

In each of the Figures~\ref{fig:gamma-01}-\ref{fig:gamma-0005}, we make the following observations: (i) SGD results in reasonably slow convergence. This observation is in line with other SGD methods using $1/L$ as the step-size~\cite{schmidt2017minimizing}. (ii) Acc-SGD with $\eta = \tau/L$ is consistently stable and as suggested by the theory, it results in faster convergence as compared to using SGD. (iii) For larger values of $\tau$ (Figures~\ref{fig:gamma-01}-~\ref{fig:gamma-005}), the training loss becomes equal to zero, verifying the interpolation property. 

The next set of experiments (Figure~\ref{fig:real}) considers binary classification on the CovType\footnote{\protect{\url{http://osmot.cs.cornell.edu/kddcup}}} and Protein\footnote{\protect{\url{http://www.csie.ntu.edu.tw/~cjlin/libsvmtools/datasets}}} datasets. For this, we train a linear classifer using the radial basis (non-parametric) features. Non-parametric regression models of this form are capable of interpolating the data~\cite{ma2018power} and thus satisfy our assumptions. We subsample $n = 8000$ random points from the datasets and use the squared-hinge loss as above. In this case, we perform a grid-search to obtain a good estimate of $\rho$. We choose $\rho = 1$ for the CovType dataset and equal to $0.1$ for the Protein dataset.

From Figures~\ref{fig:covtype} and~\ref{fig:protein}, we make the following observations: (i) In Figure~\ref{fig:covtype}, both variants have similar performance. (ii) In Figure~\ref{fig:protein}, the Acc-SGD leads to considerably faster convergence as compared to SGD. These experiments show that in cases where the interpolation property is satisfied, both SGD and accelerated SGD with a constant step-size can result in good empirical performance. 
\section{Conclusion}
\label{sec:conclusion}
In this paper, we showed that under interpolation, the stochastic gradients of common loss functions satisfy specific growth conditions. Under these conditions, we proved that it is possible for constant step-size SGD (with and without Nesterov acceleration) to achieve the convergence rates of the corresponding deterministic settings. These are the first results achieving optimal rates in the accelerated and non-convex settings under interpolation-like conditions. We used these results to demonstrate the fast convergence of the stochastic perceptron algorithm employing the squared-hinge loss. We showed that both SGD and accelerated SGD with a constant step-size can lead to good empirical performance when the interpolation property is satisfied. As opposed to determining the step-size and the schedule for annealing it for current SGD-like methods, our results imply that under interpolation, we only need to automatically determine the constant step-size for SGD. In the future, we hope to develop line-search techniques for automatically determining this step-size for both the accelerated and non-accelerated variants. 

\newpage

\section{Acknowledgements}
\label{sec:ack}
We acknowledge support from the European Research Council (grant SEQUOIA 724063) and the CIFAR program on Learning with Machines and Brains. We also thank Nicolas Flammarion, Reza Babanezhad and Adrien Taylor for discussions related to this work. We also thank Kevin Scaman for discussions and insights on using acceleration with multiplicative noise.
\bibliographystyle{plain}
\bibliography{ref}

\clearpage

\onecolumn
\appendix
\section{Incorporating additive error for Nesterov acceleration}
\label{app:additive}
For this section, we assume an additive error in the the strong growth condition implying that the following equation is satisfied for all $\x$, $z$. 
\begin{align*}
\E_{z} \normsq{\sgrad{\x}{z}} & \leq \rho \normsq{\grad{\x}} + \sigma^2
\end{align*}
In this case, we have the counterparts of Theorems~\ref{thm:sc-a} and~\ref{thm:c-a} as follows:
\begin{theorem}[Strongly convex]
Under $L$-smoothness and $\mu$ strongly-convexity, if $f$ satisfies SGC with constant $\rho$ and an additive error $\sigma$, then SGD with Nesterov acceleration with the following choice of parameters,  
\begin{flalign*}
\gk  & = \frac{1}{\sqrt{\mu \eta \rho}} \quad \text{;} \quad \bgk = 1 - \sqrt{\frac{\mu \eta}{\rho}} \\
\bkk & =  \frac{\sqrt{\mu}}{\left( 1 - \sqrt{\frac{\mu \eta}{\rho}} \right)^{(k+1)/2}} \\
\akk & = \frac{1}{\left( 1 - \sqrt{\frac{\mu \eta}{\rho}} \right)^{(k+1)/2}} \\ 
\agk & = \frac{\gk \bgk \bkk^2 \eta}{\gk \bgk \bkk^2 \eta + \ak^2} \text{;} \quad \eta = \frac{1}{\rho L} 
\end{flalign*}
results in the following convergence rate:
\begin{flalign*}
\left[ \E[f(\xkk)] - f(\xopt) \right] & \leq \left( 1 - \sqrt{\frac{\mu \eta}{\rho}} \right)^{k} \left[ f(x_0) - f(\xopt)  + \frac{\mu}{2} \normsq{x_0 - \xopt} \right] + \frac{\sigma^2 \sqrt{\eta}}{\sqrt{\rho \mu}}
\end{flalign*}
\label{thm:sc-a-additive}
\end{theorem}

\begin{theorem}[Convex]
Under $L$-smoothness and convexity, if $f$ satisfies SGC with constant $\rho$ and an additive error $\sigma$, then SGD with Nesterov acceleration with the following choice of parameters,  
\begin{flalign*}
\gk  & = \frac{\frac{1}{\rho} + \sqrt{\frac{1}{\rho^2} + 4 \gamma_{k-1}^2}}{2} \\
\akk & = \gk \sqrt{\eta \rho} \\
\agk & = \frac{\gk \eta}{\gk \eta + \ak^2} \text{;} \quad \eta = \frac{1}{\rho L} 
\end{flalign*}
results in the following convergence rate:
\begin{flalign*}
\left[ \E f(\xkk) - f(\xopt) \right] & \leq \frac{2 \rho}{k^2 \eta} \normsq{x_0 - \xopt} + \frac{k \sigma^2 \eta}{\rho} 
\end{flalign*}
\label{thm:c-a-additive}
\end{theorem}

The above theorems are proved in appendices~\ref{app:sc-a} and~\ref{app:c-a}
\section{Proofs}
\label{app:proofs}
\subsection{Proofs for SGD with Nesterov Acceleration}
Recall the update equations for SGD with Nesterov acceleration as follows:
\begin{align*}
\xkk &= \yk - \eta \sgrad{\yk}{\zk} \\
\yk  &= \agk \vk + (1 - \agk)\xk  \\
\vkk &= \bgk \vk + (1 - \bgk) \yk - \gk \eta \sgrad{\yk}{\zk} 
\end{align*}
\begin{align}
\intertext{Since the stochastic gradients are unbiased, we obtain the following equation,}
\E_{z} \left[ \sgrad{y}{z} \right] &= \grad{y} & \label{eq:unbias} \\
\intertext{For the proof, we consider the more general strong-growth condition with an additive error $\sigma^2$. }
\E_{z} \normsq{\sgrad{\x}{z}} & \leq \rho \normsq{\grad{\x}} + \sigma^2 & \label{eq:sg}
\intertext{We choose the parameters $\gk$, $\agk$, $\bgk$, $\ak$, $\bk$ such that the following equations are satisfied:}
\gk  &= \frac{1}{\rho} \cdot \left[1 + \frac{\bgk (1 - \agk)}{\agk} \right] \label{eq:gamma-update} \\
\agk &= \frac{\gk \bgk \bkk^2 \eta}{\gk \bgk \bkk^2 \eta + \ak^2} \label{eq:alpha-update} \\	
\bgk & \geq  1 - \gk \mu \eta \label{eq:beta-update} \\
\akk &= \gk \sqrt{\eta \rho} \bkk \label{eq:a-update} \\
\bkk & \leq \frac{\bk}{\sqrt{\bgk}} \label{eq:b-update} 
\end{align}

We now prove the following lemma assuming that the function $f(\cdot)$ is $L$-smooth and $\mu$ strongly-convex. 
\begin{lemma}
Assume that the function is $L$-smooth and $\mu$ strongly-convex and satisfies the strong-growth condition in Equation~\ref{eq:sg}. Then, using the updates in Equation~\ref{eq:grad-step}-\ref{eq:egrad-step} and setting the parameters according to Equations~\ref{eq:gamma-update}-~\ref{eq:b-update}, if $\eta \leq \frac{1}{\rho L}$, then the following relation holds:
\begin{align*}
\bkk^2 \gk^2 \left[ \E f(\xkk) - f^* \right] & \leq \frac{a_0^2}{\rho \eta} \left[ f(x_0) - f^* \right] + \frac{b_0^2}{2 \rho \eta} \normsq{x_0 - \xopt} + \frac{\sigma^2 \eta}{\rho} \sum_{i = 0}^{k} [ \gamma_i^2 b_{i+1}^2] 
\end{align*}
\begin{proof}
\begin{align*}
\intertext{Let $\rkk = \norm{\vkk - \xopt}$, then using equation~\ref{eq:egrad-step}}
& \rkk^2 = \normsq{\bgk \vk + (1 - \bgk) \yk - \xopt - \gk \eta \sgrad{\yk}{\zk}} \\
\rkk^2 & = \normsq{\bgk \vk + (1 - \bgk) \yk - \xopt} + \gk^2 \eta^2 \normsq{\sgrad{\yk}{\zk}} + 2 \gk \eta \langle \xopt - \bgk \vk - (1 - \bgk) \yk, \sgrad{\yk}{\zk} \rangle 
\end{align*}
\begin{align*}
\intertext{Taking expecation wrt to $\zk$,}
& \E[\rkk^2] = \E[\normsq{\bgk \vk + (1 - \bgk) \yk - \xopt}] + \gk^2 \eta^2 \E \normsq{\sgrad{\yk}{\zk}} + 2 \gk \eta \left[ \E \langle \xopt - \bgk \vk - (1 - \bgk) \yk, \sgrad{\yk}{\zk} \rangle \right] \\
& \leq \normsq{\bgk \vk + (1 - \bgk) \yk - \xopt} + \gk^2  \eta^2 \rho \normsq{\grad{\yk}} + 2 \gk \eta \left[ \langle \xopt - \bgk \vk - (1 - \bgk) \yk, \grad{\yk} \rangle \right] + \gk^2 \eta^2 \sigma^2 \\
& = \normsq{\bgk (\vk - \xopt) + (1 - \bgk) (\yk - \xopt)} + \gk^2 \eta^2 \rho \normsq{\grad{\yk}} + 2 \gk  \eta \left[ \langle \xopt - \bgk \vk - (1 - \bgk) \yk, \grad{\yk} \rangle \right] + \gk^2 \eta^2 \sigma^2  \\
& \leq \bgk \normsq{\vk - \xopt} + (1 - \bgk) \normsq{\yk - \xopt} + \gk^2 \eta^2 \rho \normsq{\grad{\yk}} + 2 \gk \eta \left[ \langle \xopt - \bgk \vk - (1 - \bgk) \yk, \grad{\yk} \rangle \right] + \gk^2 \eta^2 \sigma^2   \tag{By convexity of $\normsq{\cdot}$}\\
& = \bgk \rk^2 + (1 - \bgk) \normsq{\yk - \xopt} + \gk^2 \eta^2 \rho \normsq{\grad{\yk}} + 2 \gk \eta \left[ \langle \xopt - \bgk \vk - (1 - \bgk) \yk, \grad{\yk} \rangle \right] + \gk^2 \eta^2 \sigma^2  \\
& = \bgk \rk^2 + (1 - \bgk) \normsq{\yk - \xopt} + \gk^2  \eta^2 \rho \normsq{\grad{\yk}} + 2 \gk  \eta \left[ \langle \bgk (\yk - \vk) + \xopt - \yk, \grad{\yk} \rangle \right] + \gk^2 \eta^2 \sigma^2  \\
& = \bgk \rk^2 + (1 - \bgk) \normsq{\yk - \xopt} + \gk^2  \eta^2 \rho \normsq{\grad{\yk}} + 2 \gk  \eta \left[ \langle \frac{\bgk (1 - \agk)}{\agk} \left( \xk - \yk \right) + \xopt - \yk, \grad{\yk} \rangle \right] + \gk^2 \eta^2 \sigma^2 \tag{From equation~\ref{eq:mix-step}} \\
& = \bgk \rk^2 + (1 - \bgk) \normsq{\yk - \xopt} + \gk^2 \eta^2 \rho \normsq{\grad{\yk}} + 2 \gk \eta \left[ \frac{\bgk (1 - \agk)}{\agk} \langle \grad{\yk}, \left( \xk - \yk \right) \rangle + \langle \grad{\yk}, \xopt - \yk \rangle \right] + \gk^2 \eta^2 \sigma^2 \\
& \leq \bgk \rk^2 + (1 - \bgk) \normsq{\yk - \xopt} + \gk^2 \eta^2 \rho \normsq{\grad{\yk}} + 2 \gk \eta \left[ \frac{\bgk (1 - \agk)}{\agk} \left( f(\xk) - f(\yk) \right) + \langle \grad{\yk}, \xopt - \yk \rangle\right] + \gk^2 \eta^2 \sigma^2 \tag{By convexity} 
\end{align*}
\begin{align}
\intertext{By strong convexity,} 
& \E[\rkk^2] \leq \bgk \rk^2 + (1 - \bgk) \normsq{\yk - \xopt} + \gk^2 \eta^2 \rho \normsq{\grad{\yk}} \nonumber \\ 
& + 2 \gk \eta \left[ \frac{\bgk (1 - \agk)}{\agk} \left( f(\xk) - f(\yk) \right) + f^* - f(\yk) - \frac{\mu}{2} \normsq{\yk - \xopt} \right]  + \gk^2 \eta^2 \sigma^2  \label{eq:inter}
\intertext{By Lipschitz continuity of the gradient,}
& f(\xkk) - f(\yk)  \leq \langle \grad{\yk}, \xkk - \yk \rangle + \frac{L}{2} \normsq{\xkk - \yk} \nonumber \\
& \leq - \eta \langle \grad{\yk}, \sgrad{\yk}{\zk} \rangle + \frac{L \eta^2}{2} \normsq{\sgrad{\yk}{\zk}} \nonumber \\
\intertext{Taking expectation wrt $\zk$ and using equations~\ref{eq:unbias},~\ref{eq:sg}}
& \E[f(\xkk) - f(\yk)] \leq - \eta \normsq{\grad{\yk}} + \frac{L \rho \eta^2}{2} \normsq{\grad{\yk}} + \frac{L \eta^2 \sigma^2}{2} \nonumber \\
& \E[f(\xkk) - f(\yk)] \leq  \left[ - \eta + \frac{L \rho \eta^2}{2} \right] \normsq{\grad{\yk}} + \frac{L \eta^2 \sigma^2}{2} \nonumber \\
\intertext{If $\eta \leq \frac{1}{\rho L}$,}
& \E[f(\xkk) - f(\yk)] \leq  \left(\frac{-\eta}{2}\right) \normsq{\grad{\yk}} + \frac{L \eta^2 \sigma^2}{2} \nonumber  \\
& \implies \normsq{\grad{\yk}} \leq \left(\frac{2}{\eta}\right) \E[f(\yk) - f(\xkk)] + L \eta \sigma^2 \label{eq:descent-lemma}
\end{align}
\begin{align*}
\intertext{From equations~\ref{eq:inter} and~\ref{eq:descent-lemma},}
\E[\rkk^2] & \leq \bgk \rk^2 + (1 - \bgk) \normsq{\yk - \xopt} + 2 \gk^2 \rho \eta \E[f(\yk) - f(\xkk)] \\ & + 2 \gk \eta \left[ \frac{\bgk (1 - \agk)}{\agk} \left( f(\xk) - f(\yk) \right) + f^* - f(\yk) - \frac{\mu}{2} \normsq{\yk - \xopt} \right] + \gk^2 \eta^2 \sigma^2 + L \gk^2 \eta^3 \rho \sigma^2 \\
& \leq \bgk \rk^2 + (1 - \bgk) \normsq{\yk - \xopt} + 2 \gk^2 \eta \rho \E[f(\yk) - f(\xkk)] \\ & + 2 \gk \eta \left[ \frac{\bgk (1 - \agk)}{\agk} \left( f(\xk) - f(\yk) \right) + f^* - f(\yk) - \frac{\mu}{2} \normsq{\yk - \xopt} \right] + 2 \gk^2 \eta^2 \sigma^2 \tag{Since $\eta \leq \frac{1}{\rho L}$}\\
& = \bgk \rk^2 + \normsq{\yk - \xopt} \left[ (1 - \bgk) - \gk \mu \eta \right] + f(\yk) \left[2 \gk^2 \eta \rho - 2 \gk \eta \cdot \frac{\bgk (1 - \agk)}{\agk} - 2 \gk \eta \right] \\ & - 2 \gk^2 \eta \rho \E f(\xkk) + 2 \gk \eta f^* + \left[ 2 \gk \eta \cdot \frac{\bgk (1 - \agk)}{\agk} \right] f(\xk) +  2 \gk^2 \eta^2 \sigma^2 \\
\intertext{Since $\bgk \geq 1 - \gk \mu \eta$ and $\gk = \frac{1}{\rho} \cdot \left( 1 + \frac{\bgk (1 - \agk)}{\agk} \right)$,}
\E[\rkk^2] & \leq \bgk \rk^2 - 2 \gk^2 \eta \rho \E f(\xkk) + 2 \gk \eta f^* + \left[ 2 \gk \eta \cdot \frac{\bgk (1 - \agk)}{\agk} \right] f(\xk) +  2 \gk^2 \eta^2 \sigma^2 \\ 
\intertext{Multiplying by $\bkk^2$,}
\bkk^2 \E[\rkk^2] & \leq \bkk^2 \bgk \rk^2 - 2 \bkk^2 \gk^2 \eta \rho \E f(\xkk) + 2 \bkk^2 \gk \eta f^* + \left[ 2 \bkk^2 \gk \eta \cdot \frac{\bgk (1 - \agk)}{\agk} \right] f(\xk) +  2 \bkk^2 \gk^2 \eta^2 \sigma^2 \\
\intertext{Since $\bkk^2 \bgk \leq \bk^2$, $\bkk^2 \gk^2 \eta \rho = \akk^2$, $\frac{\gk \eta \bgk (1 - \agk)}{\agk} = \frac{\ak^2}{\bkk^2}$}
\bkk^2 \E[\rkk^2] & \leq \bk^2 \rk^2 - 2 \akk^2 \E f(\xkk) + 2 \bkk^2 \gk \eta f^* + 2 \ak^2 f(\xk) + \frac{2 \akk^2 \sigma^2 \eta}{\rho} \\
& = \bk^2 \rk^2 - 2 \akk^2 \left[ \E f(\xkk) - f^* \right] + 2 \ak^2 \left[ f(\xk) - f^* \right]  + 2 \left[\bkk^2 \gk \eta - \akk^2 + \ak^2 \right] f^*  + \frac{2 \akk^2 \sigma^2 \eta}{\rho} \\
\intertext{Since $\left[\bkk^2 \gk \eta - \akk^2 + \ak^2 \right] = 0$,}
\bkk^2 \E[\rkk^2] & \leq \bk^2 \rk^2 - 2 \akk^2 \left[ \E f(\xkk) - f^* \right] + 2 \ak^2 \left[ f(\xk) - f^* \right] + \frac{2 \akk^2 \sigma^2 \eta}{\rho}
\end{align*}
\begin{align}
\intertext{Denoting $\E f(\xkk)$ as $\phi_k$,}
2 \akk^2 \left[ \phi_{k+1} - f^* \right] + \bkk^2 \E[\rkk^2] & \leq 2 \ak^2 \left[ \phi_{k} - f^* \right] + \bk^2 \E[\rk^2] + \frac{2 \akk^2 \sigma^2 \eta}{\rho} \nonumber \\
\intertext{By recursion,}
2 \akk^2 \left[ \phi_{k+1} - f^* \right] + \bkk^2 \E[\rkk^2] & \leq 2 a_0^2 \left[ f(x_0) - f^* \right] + b_0^2 \normsq{x_0 - \xopt} + \frac{2 \sigma^2 \eta}{\rho} \sum_{i = 0}^{k} [ a_{i+1}^2 ] \nonumber \\
2 \akk^2 \left[ \phi_{k+1} - f^* \right] & \leq 2 a_0^2 \left[ f(x_0) - f^* \right] + b_0^2 \normsq{x_0 - \xopt} + \frac{2 \sigma^2 \eta}{\rho} \sum_{i = 0}^{k} [ a_{i+1}^2 ] \nonumber \\
2 \bkk^2 \gk^2 \rho \eta \left[ \phi_{k+1} - f^* \right] & \leq 2 a_0^2 \left[ f(x_0) - f^* \right] + b_0^2 \normsq{x_0 - \xopt} + 2 \sigma^2 \eta^2 \rho \sum_{i = 0}^{k} [ \gamma_i^2 b_{i+1}^2 ] \nonumber \\
\bkk^2 \gk^2 \left[ \E f(\xkk) - f^* \right] & \leq \frac{a_0^2}{\rho \eta} \left[ f(x_0) - f^* \right] + \frac{b_0^2}{2 \rho \eta} \normsq{x_0 - \xopt} + \frac{\sigma^2 \eta}{\rho} \sum_{i = 0}^{k} [ \gamma_i^2 b_{i+1}^2] & \nonumber
\end{align}
\end{proof}
\label{lemma:inter}
\end{lemma}

\begin{lemma}
Under the parameter setting according to Equations~\ref{eq:gamma-update}-~\ref{eq:b-update}, the following relation is true:
\begin{align*}
\gk^2 - \gk \left[ \frac{1}{\rho} -  \mu \eta \gamma^2_{k-1} \right] = \gamma^2_{k-1}
\end{align*}
\begin{proof}
\begin{align}
\gk & = \frac{1}{\rho} \left[ 1 + \frac{\bgk (1 - \agk)}{\agk} \right] & \tag{From equation~\ref{eq:gamma-update}} \nonumber \\
\gk^2 - \frac{\gk}{\rho} & = \frac{\gk \bgk (1 - \agk)}{\rho \agk} \nonumber \\
& = \frac{1}{\eta \rho} \frac{\ak^2}{\bkk^2}  & \tag{From equation~\ref{eq:alpha-update}} \nonumber \\
& = \frac{\bgk}{\eta \rho} \frac{\ak^2}{\bk^2} & \tag{From equation~\ref{eq:b-update}} \nonumber \\
& = \frac{1 - \gk \mu \eta}{\eta \rho} \frac{\ak^2}{\bk^2} & \tag{From equation~\ref{eq:beta-update}} \nonumber \\
& = \frac{1 - \gk \mu \eta}{\eta \rho} \left( \gamma_{k-1} \sqrt{\eta \rho} \right)^2  & \tag{From equation~\ref{eq:beta-update}} \nonumber \\
& = \left( 1 - \gk \mu \eta \right) \gamma^2_{k-1} \nonumber  \\
\implies \gk^2 - \gk \left[ \frac{1}{\rho} -  \mu \eta \gamma^2_{k-1} \right] = \gamma^2_{k-1} 
\end{align}
\end{proof}
\label{lemma:gamma}
\end{lemma}
\subsubsection{Strongly-convex case}
\label{app:sc-a}
We now consider the strongly-convex case, 
\begin{align*}
\intertext{Using Lemma~\ref{lemma:gamma},}
\gk^2 - \gk \left[ \frac{1}{\rho} -  \mu \eta \gamma^2_{k-1} \right] & = \gamma^2_{k-1} \\
\intertext{If $\gk = C$, then}
\gk & = \frac{1}{\sqrt{\mu \eta \rho}} \\
\bgk & = 1 - \sqrt{\frac{\mu \eta}{\rho}} \\
\bkk &=  \frac{b_0}{\left( 1 - \sqrt{\frac{\mu \eta}{\rho}} \right)^{(k+1)/2}} \\
\akk &= \frac{1}{\sqrt{\mu \eta \rho}} \cdot \sqrt{\eta \rho} \cdot \frac{b_0}{\left( 1 - \sqrt{\frac{\mu \eta}{\rho}} \right)^{(k+1)/2}} = \frac{b_0}{\sqrt{\mu}} \cdot \frac{1}{\left( 1 - \sqrt{\frac{\mu \eta}{\rho}} \right)^{(k+1)/2}} \\
\intertext{If $b_0 = \sqrt{\mu}$,}
\akk &= \frac{1}{\left( 1 - \sqrt{\frac{\mu \eta}{\rho}} \right)^{(k+1)/2}} \\
\intertext{The above equation implies that $a_0 = 1$. This gives us the parameter settings used in Theorem~\ref{thm:sc-a}.}
\end{align*}
\begin{align*}
\intertext{Using the result of Lemma~\ref{lemma:inter} and the above relations, we obtain the following inequality. Note that $\phi_{k+1} = \E[f(\xkk)]$.}
\frac{\mu}{\left( 1 - \sqrt{\frac{\mu \eta}{\rho}} \right)^{(k+1)}} \cdot \frac{1}{\mu \eta \rho} \left[ \phi_{k+1} - f^* \right] & \leq \frac{1}{\rho \eta} \left[ f(x_0) - f^* \right] + \frac{\mu}{2 \rho \eta} \normsq{x_0 - \xopt} + \frac{\sigma^2 \eta}{\rho} \cdot \frac{1}{\mu \eta \rho} \sum_{i = 0}^{k} \frac{\mu}{\left( 1 - \sqrt{\frac{\mu \eta}{\rho}} \right)^{(i+1)}} & \\
\frac{1}{\left( 1 - \sqrt{\frac{\mu \eta}{\rho}} \right)^{k}} \left[ \phi_{k+1} - f^* \right] & \leq \left[ f(x_0) - f^* \right] + \frac{\mu}{2} \normsq{x_0 - \xopt} + \frac{\sigma^2 \eta}{\rho} \sum_{i = 0}^{k} \frac{1}{\left( 1 - \sqrt{\frac{\mu \eta}{\rho}} \right)^{(i+1)}} & \\
\frac{1}{\left( 1 - \sqrt{\frac{\mu \eta}{\rho}} \right)^{k}} \left[ \phi_{k+1} - f^* \right] & \leq \left[ f(x_0) - f^*  + \frac{\mu}{2} \normsq{x_0 - \xopt} \right] + \frac{\sigma^2 \sqrt{\eta}}{\sqrt{\rho \mu}} \left( 1 - \sqrt{\frac{\mu \eta}{\rho}} \right)^{-k}   \\
\left[ \phi_{k+1} - f^* \right] & \leq \left( 1 - \sqrt{\frac{\mu \eta}{\rho}} \right)^{k} \left[ f(x_0) - f^*  + \frac{\mu}{2} \normsq{x_0 - \xopt} \right] + \frac{\sigma^2 \sqrt{\eta}}{\sqrt{\rho \mu}}
\end{align*}
\subsubsection{Proof of Theorem~\ref{thm:sc-a}}
\begin{align*}
\intertext{We use the above relation to complete the proof for Theorem~\ref{thm:sc-a}. Substituting $\eta = \frac{1}{\rho L}$ and $\sigma = 0$, we obtain the following:}
\left[ \E[f(\xkk)] - f^* \right] & \leq \left( 1 - \sqrt{\frac{\mu \eta}{\rho}} \right)^{k} \left[ f(x_0) - f^*  + \frac{\mu}{2} \normsq{x_0 - \xopt} \right] 
\end{align*}
\subsubsection{Convex case}
\label{app:c-a}
We now use the above lemmas to first prove the convergence rate in the convex case. In this case, $\mu = 0$ and the result of Lemma~\ref{lemma:gamma} can be written as:
\begin{align*}
\gk^2 - \frac{\gk}{\rho}- \gamma^2_{k-1} & = 0 \\
\implies \gk & = \frac{\frac{1}{\rho} + \sqrt{\frac{1}{\rho^2} + 4 \gamma_{k-1}^2}}{2} \\
\intertext{Let $\gamma_0 = 0$. From equation~\ref{eq:beta-update}, for all $k$,}
\bgk & = 1 \\
\bkk & = \bk = b_0 = 1 & \tag{From equation~\ref{eq:b-update}} \\
\akk &= \gk \sqrt{\eta \rho} b_0 \implies \akk = \gk \sqrt{\eta \rho} & \tag{From equation~\ref{eq:a-update}} \\
\intertext{The above equation implies that $a_0 = 0$. This gives us the parameter settings used in Theorem~\ref{thm:c-a}.}
\end{align*}
\begin{align*}
\intertext{Using the result of Lemma~\ref{lemma:inter} by setting $\mu = 0$ and the above relations, we obtain the following inequality. Note that $\phi_{k+1} = \E[f(\xkk)]$.}
\gk^2 \left[ \phi_{k+1} - f^* \right] & \leq \frac{1}{2 \rho \eta} \normsq{x_0 - \xopt} + \frac{\sigma^2 \eta}{\rho} \sum_{i = 1}^{k-1} [ \gamma_i^2 ] & \\
\intertext{By induction, $\gamma_i \geq \frac{i}{2 \rho}$,}
\frac{k^2}{4 \rho^2} \left[ \phi_{k+1} - f^* \right] & \leq \frac{1}{2 \rho \eta} \normsq{x_0 - \xopt} + \frac{\sigma^2 \eta}{4 \rho^3} \sum_{i = 1}^{k-1} [ i^2 ] \\
\left[ \phi_{k+1} - f^* \right] & \leq \frac{2 \rho}{k^2 \eta} \normsq{x_0 - \xopt} + \frac{\sigma^2 \eta}{k^2 \rho} \sum_{i = 1}^{k-1} [ i^2 ] \\
\left[ \phi_{k+1} - f^* \right] & \leq \frac{2 \rho}{k^2 \eta} \normsq{x_0 - \xopt} + \frac{k \sigma^2 \eta}{\rho} 
\end{align*}
\subsubsection{Proof of Theorem~\ref{thm:c-a}}
\begin{align*}
\intertext{We use the above relation to complete the proof for Theorem~\ref{thm:c-a}. Substituting $\eta = \frac{1}{\rho L}$ and $\sigma = 0$, we obtain the following:}
\left[ \E[f(\xkk)] - f^* \right] & \leq \frac{2 \rho^2 L}{k^2} \normsq{x_0 - \xopt}
\end{align*}
\subsection{Proof of Theorem~\ref{thm:nc-na}}
\label{app:nc-na}
\begin{proof}
Recall the stochastic gradient descent update, 
\begin{align}
& \xkk = \xk - \eta \sgrad{\xk}{\zk} & \label{eq:nc-update} \\
\intertext{By Lipschitz continuity of the gradient,}
& f(\xkk) - f(\xk)  \leq \langle \grad{\xk}, \xkk - \xk \rangle + \frac{L}{2} \normsq{\xkk - \xk} \nonumber \\
& \leq - \eta \langle \grad{\xk}, \sgrad{\xk}{\zk} \rangle + \frac{L \eta^2}{2} \normsq{\sgrad{\xk}{\zk}} \nonumber \\
\intertext{Taking expectation wrt $\zk$ and using equations~\ref{eq:unbias},~\ref{eq:sg}}
& \E[f(\xkk) - f(\xk)] \leq - \eta \normsq{\grad{\xk}} + \frac{L \rho \eta^2}{2} \normsq{\grad{\xk}} + \frac{L \eta^2 \sigma^2}{2} \nonumber \\
& \E[f(\xkk) - f(\xk)] \leq  \left[ - \eta + \frac{L \rho \eta^2}{2} \right] \normsq{\grad{\xk}} + \frac{L \eta^2 \sigma^2}{2} \nonumber \\
\intertext{If $\eta \leq \frac{1}{\rho L}$,}
& \E[f(\xkk) - f(\xk)] \leq  \left(\frac{-\eta}{2}\right) \normsq{\grad{\xk}} + \frac{L \eta^2 \sigma^2}{2} \nonumber  \\
& \implies \normsq{\grad{\xk}} \leq \left(\frac{2}{\eta}\right) \E[f(\xk) - f(\xkk)] + L \eta \sigma^2 & \label{eq:nc-descent-lemma}
\intertext{Taking expectation wrt $z_{0}, z_{1}, \ldots z_{t-1}$ and summing from $k = 0$ to $t-1$,}
& \sum_{k = 0}^{t-1} \E \left[ \normsq{\grad{\xk}} \right]  \leq \left(\frac{2}{\eta}\right) \sum_{k = 0}^{t-1} \E \left[ f(\xk) - f(\xkk)\right] + L \eta t \sigma^2  & \nonumber \\
& \implies \sum_{k = 0}^{t-1} \min_{k = 0,1, \ldots t-1} \E \left[ \normsq{\grad{\xk}} \right] \leq \left(\frac{2}{\eta}\right) \sum_{k = 0}^{t-1} \E \left[ f(\xk) - f(\xkk)\right] + L \eta \sigma^2 & \nonumber \\
& \min_{k = 0,1, \ldots t-1} \E \left[ \normsq{\grad{\xk}} \right] \leq \left(\frac{2}{\eta t}\right) \left[ f(\x_{0}) - \E[f(\x_{t})] \right] + L \eta \sigma^2 & \nonumber \\
& \min_{k = 0,1, \ldots t-1} \E \left[ \normsq{\grad{\xk}} \right] \leq \left(\frac{2}{\eta t}\right) \left[ f(\x_{0}) - f(\xopt)\right] + L \eta \sigma^2 & \nonumber 
\intertext{If $\sigma = 0$,}
& \min_{k = 0,1, \ldots t-1} \E  \left[ \normsq{\grad{\xk}} \right] \leq \left(\frac{2}{\eta t}\right) \left[ f(\x_{0}) - f(\xopt)\right] & \nonumber \\
\implies & \min_{k = 0,1, \ldots t-1} \E  \left[ \normsq{\grad{\xk}} \right] \leq \left(\frac{2 \rho L}{t} \right) \left[ f(\x_{0}) - f(\xopt) \right] & \tag{Setting $\eta = \frac{1}{\rho L}$} 
\end{align}
\end{proof}

\subsection{Proof of Theorem~\ref{thm:nc-pl-na}}
\label{app:nc-pl-na}
\begin{proof}
Similar to the proof of Theorem~\ref{thm:nc-na}, we can use the SGD update and Lipschitz continuity of the gradient to obtain the following equation for the stepsize $\eta \leq \frac{1}{\rho L}$:
\begin{align}
& \E[f(\xkk) - f(\xk)] \leq  \left(\frac{-\eta}{2}\right) \normsq{\grad{\xk}} + \frac{L \eta^2 \sigma^2}{2} \nonumber  \\
\intertext{We now use the PL inequality with constant $\mu$ as follows:}
& \normsq{\grad{\xk}} \geq 2 \mu \left[ f(\xk) - f^* \right] \nonumber \\
\intertext{Combining the above two inequalities,}
& \E[f(\xkk) - f(\xk)] \leq  -\eta \mu \left[ f(\xk) - f^* \right] + \frac{L \eta^2 \sigma^2}{2} \nonumber  \\
\intertext{If $\sigma = 0$,}
& \E[f(\xkk) - f(\xk)] \leq  -\eta \mu \left[ f(\xk) - f^* \right] \nonumber  \\
\implies & \E[f(\xkk) - f^*] \leq  \left( 1 -\eta \mu \right) \left[ f(\xk) - f^* \right] \nonumber  \\
\intertext{Substituting $\eta = \frac{1}{\rho L}$,}
& \E[f(\xkk) - f^*] \leq  \left( 1 - \frac{\mu}{\rho L} \right) \left[ f(\xk) - f^* \right] \nonumber  \\
\implies & \E[f(\xkk) - f^*] \leq  \left( 1 - \frac{\mu}{\rho L} \right)^{k} \left[ f(\x_{0}) - f^* \right] \nonumber  \\
\end{align}
\end{proof}

\subsection{Proof of Theorem~\ref{thm:sc-na}}
\label{app:sc-na}
\begin{proof}
\begin{align*}
\normsq{\xkk - \xopt} & = \normsq{\xk - \eta \sgrad{\xk}{z} - \xopt} & \\
& = \normsq{\xk - \xopt} - 2 \eta \langle \sgrad{\xk}{z}, \xk - \xopt \rangle + \eta^2 \normsq{\sgrad{\xk}{z}} \\
\E_z [\normsq{\xkk - \xopt}] & = \normsq{\xk - \xopt} - 2 \eta \E[\langle \sgrad{\xk}{z}, \xk - \xopt  \rangle] + \eta^2 \E[\normsq{\sgrad{\xk}{z}}] \\
& = \normsq{\xk - \xopt} - 2 \eta \langle \grad{\xk}, \xk - \xopt  \rangle + \eta^2 \E[\normsq{\sgrad{\xk}{z}}] & \tag{From the unbiasedness of stochastic gradients.} \\
& \leq \normsq{\xk - \xopt} - 2 \eta \langle \grad{\xk}, \xk - \xopt  \rangle + 2 \rho \eta^2 L [f(\xk) - f^*] & \tag{From equation~\ref{eq:wgc-main}} \\
& \leq \normsq{\xk - \xopt} + 2 \eta \left[ f^* - f(\xk) - \frac{\mu}{2} \normsq{\xk - \xopt} \right] + 2 \rho \eta^2 L [f(\xk) - f^*] & \tag{By strong convexity} \\
& = \left(1 - \mu \eta \right) \normsq{\xk - \xopt} +  \left( 2 \eta^2 \rho L - 2 \eta \right) \left[ f(\xk) - f^* \right] \\
\normsq{\xkk - \xopt} & \leq \left(1 - \frac{\mu}{\rho L} \right) \normsq{\xk - \xopt} & \tag{Setting $\eta = \frac{1}{\rho L}$} \\
\implies \normsq{\xkk - \xopt} & \leq \left(1 - \frac{\mu}{\rho L} \right)^{k} \normsq{x_0 - \xopt}
\end{align*}
\end{proof}

\subsection{Proof of Theorem~\ref{thm:c-na}}
\label{app:c-na}
\begin{proof}
\begin{align*}
\intertext{By convexity,}
f(\xk) & \leq f(\xopt) + \langle \grad{\xk}, \xk - \xopt \rangle \\
\intertext{For any $\beta \leq 1$,}
f(\xk) & \leq \beta f(\xk) + (1 - \beta) f(\xopt) + (1 - \beta) \langle \grad{\xk}, \xk - \xopt  \rangle \\
\intertext{By Lipschitz continuity of $\grad{f}$,}
f(\xkk) & \leq f(\xk) + \langle \grad{\xk}, \xkk - \xk \rangle + \frac{L}{2} \normsq{\xkk - \xk} \\
\implies f(\xkk) & \leq f(\xk) -  \eta \langle \grad{\xk}, \sgrad{\xk}{z} \rangle + \frac{\eta^2 L}{2} \normsq{\sgrad{\xk}{z}} & \\
\intertext{From the above equations,}
f(\xkk) & \leq \beta f(\xk) + (1 - \beta) f(\xopt) + (1 - \beta) \langle \grad{\xk}, \xk - \xopt  \rangle  - \eta \langle \grad{\xk}, \sgrad{\xk}{z} \rangle + \frac{\eta^2 L}{2} \normsq{\sgrad{\xk}{z}} 
\end{align*}
Note that,
\begin{align*}
\frac{1}{2 \eta} \left( \normsq{\xk - \xopt} - \normsq{\xkk - \xopt} \right) & = \frac{1}{2 \eta} \left( \normsq{\xk - \xopt} - \normsq{\xk - \eta \sgrad{\xk}{z} - \xopt} \right) \\
& = \frac{1}{2 \eta} \left( \normsq{\xk - \xopt} - \normsq{\xk - \xopt} - \eta^2 \normsq{\sgrad{\xk}{z}} + 2 \eta \langle \xk - \xopt, \sgrad{\xk}{z} \rangle  \right) \\
\frac{1}{2 \eta} \left( \normsq{\xk - \xopt} - \normsq{\xkk - \xopt} \right) & = \frac{-\eta}{2} \normsq{\sgrad{\xk}{z}} + \langle \xk - \xopt, \sgrad{\xk}{z} \rangle \\
\implies \langle \xk - \xopt, \sgrad{\xk}{z} \rangle &= \frac{1}{2 \eta} \left( \normsq{\xk - \xopt} - \normsq{\xkk - \xopt} \right) + \frac{\eta}{2} \normsq{\sgrad{\xk}{z}} \\
\intertext{Taking expectation}
\E \left[ \langle \xk - \xopt, \sgrad{\xk}{z} \rangle \right] &= \frac{1}{2 \eta} \left( \normsq{\xk - \xopt} - \E \left[ \normsq{\xkk - \xopt} \right] \right) + \frac{\eta}{2} \E \left[ \normsq{\sgrad{\xk}{z}} \right] \\
\implies \langle \xk - \xopt, \grad{\xk} \rangle &= \frac{1}{2 \eta} \left( \normsq{\xk - \xopt} - \E \left[ \normsq{\xkk - \xopt} \right] \right) + \frac{\eta}{2} \E \left[ \normsq{\sgrad{\xk}{z}} \right] \\
\end{align*}
Using the above equations, 
\begin{align*}
f(\xkk) & \leq \beta f(\xk) + (1 - \beta) f(\xopt) + \frac{1 - \beta}{2 \eta} \left( \normsq{\xk - \xopt} - \E \left[ \normsq{\xkk - \xopt} \right] \right) + \frac{(1 - \beta)(\eta)}{2} \E \left[ \normsq{\sgrad{\xk}{z}} \right] \\ 
& - \eta \langle \grad{\xk}, \sgrad{\xk}{z} \rangle + \frac{\eta^2 L}{2} \normsq{\sgrad{\xk}{z}} \\
\intertext{Taking expectation,}
\E[f(\xkk)] & \leq \beta f(\xk) + (1 - \beta) f(\xopt) + \frac{1 - \beta}{2 \eta} \left( \normsq{\xk - \xopt} - \E \left[ \normsq{\xkk - \xopt} \right] \right) + \frac{(1 - \beta)(\eta)}{2} \E \left[ \normsq{\sgrad{\xk}{z}} \right] \\ 
& - \eta \langle \grad{\xk}, \E \left[ \sgrad{\xk}{z} \right] \rangle + \frac{\eta^2 L}{2} \E \left[ \normsq{\sgrad{\xk}{z}} \right] \\
& = \beta f(\xk) + (1 - \beta) f(\xopt) + \frac{1 - \beta}{2 \eta} \left( \normsq{\xk - \xopt} - \E \left[\normsq{\xkk - \xopt} \right] \right) + \frac{(1 - \beta)(\eta)}{2} \E \left[ \normsq{\sgrad{\xk}{z}} \right] \\ 
& - \eta \normsq{\grad{\xk}} + \frac{\eta^2 L}{2} \E \left[ \normsq{\sgrad{\xk}{z}} \right] \\
\intertext{The term $- \eta \normsq{\grad{\xk}} \leq 0$}
\implies \E[f(\xkk)] & \leq \beta f(\xk) + (1 - \beta) f(\xopt) + \frac{1 - \beta}{2 \eta} \left( \normsq{\xk - \xopt} - \E \left[\normsq{\xkk - \xopt} \right] \right) \\ 
& + \frac{(1 - \beta)(\eta)}{2} \E \left[ \normsq{\sgrad{\xk}{z}} \right] + \frac{\eta^2 L}{2} \E \left[ \normsq{\sgrad{\xk}{z}} \right] \\
\E[f(\xkk)] - f(\xopt) & \leq \beta \left( f(\xk) - f(\xopt) \right) + \frac{1 - \beta}{2 \eta} \left( \normsq{\xk - \xopt} - \E \left[\normsq{\xkk - \xopt} \right] \right) \\
& + \left( \frac{(1 - \beta)(\eta)}{2} + \frac{\eta^2 L}{2} \right) \E \left[ \normsq{\sgrad{\xk}{z}} \right] \\
\intertext{From equation~\ref{eq:wgc-main},}
\E[f(\xkk)] - f(\xopt) & \leq \beta \left( f(\xk) - f(\xopt) \right) + \frac{1 - \beta}{2 \eta} \left( \normsq{\xk - \xopt} - \E \left[\normsq{\xkk - \xopt} \right] \right) \\
& + \left( \rho (1 - \beta) \eta L + \eta^2 \rho L^2 \right) \left( f(\xk) - f(\xopt) \right) \\
\intertext{Let us choose $1 - \beta = \eta L$,}
\E[f(\xkk)] - f(\xopt) & \leq \beta \left( f(\xk) - f(\xopt) \right) + \frac{1 - \beta}{2 \eta} \left( \normsq{\xk - \xopt} - \E \left[\normsq{\xkk - \xopt} \right] \right) + 2 \rho \eta^2 L^2 \left( f(\xk) - f(\xopt) \right) \\
\E[f(\xkk)] - f(\xopt) & \leq \left( \beta + 2 \rho \eta^2 L^2 \right) \left( f(\xk) - f(\xopt) \right) + \frac{L}{2} \left( \normsq{\xk - \xopt} - \E \left[\normsq{\xkk - \xopt} \right] \right) \\
\intertext{Let $\delta_{k+1} = \E[f(\xkk)] - f(\xopt)$ and $\Delta_{k+1} = \E \left[ \normsq{\xkk - \xopt} \right]$}
\implies \delta_{k+1} & \leq \left( \beta + 2 \rho \eta^2 L^2 \right) \delta_{k} + \frac{L}{2} \left[ \Delta_{k} - \Delta_{k+1} \right] 
\end{align*}
Summing from $i = 0$ to $k - 1$,
\begin{align*}
\sum_{i = 0}^{k-1} \delta_{i+1} & \leq \left( \beta + 2 \rho \eta^2 L^2 \right) \sum_{i = 0}^{k-1} \delta_{i} + \frac{L}{2} \sum_{i = 0}^{k-1} \left[ \Delta_{i} - \Delta_{i+1} \right] \\
\implies \sum_{i = 0}^{k-1} \delta_{i+1} & \leq \left( \beta + 2 \rho \eta^2 L^2 \right) \sum_{i = 0}^{k-1} \delta_{i} + \frac{L}{2} \Delta_{0} \\
\implies \sum_{i = 1}^{k} \delta_{i} & \leq \frac{\left( \beta + 2 \rho \eta^2 L^2 \right) \delta_{0} + \frac{L}{2} \Delta_{0} }{\left(1 - \beta - 2 \rho \eta^2 L^2 \right)}
\intertext{Let $\mxk = \frac{\left[ \sum_{i = 1}^{k} \x_{i} \right]}{k}$. By Jensen's inequality,}
\E[f(\mxk)] & \leq \frac{\sum_{i = 1}^{k} \E[f(\x_{i})]}{k} \\
\implies \E[f(\mxk)] - f(\xopt) & \leq \sum_{i = 1}^{k} \delta_{i} \\
\implies \E[f(\mxk)] - f(\xopt) & \leq \frac{\left( \beta + 2 \rho \eta^2 L^2 \right) \delta_{0} + \frac{L}{2} \Delta_{0} }{\left(1 - \beta - 2 \rho \eta^2 L^2 \right) k} \\
\E[f(\mxk)]  - f(\xopt) & \leq \frac{\left( 1 - \eta L + 2 \rho \eta^2 L^2 \right) \left[f(\x_0) - f(\xopt) \right]  + \frac{L}{2} \normsq{\x_0 - \xopt} }{\left(\eta L - 2 \rho \eta^2 L^2 \right) k} & \tag{Since $1 - \beta = \eta L$}
\intertext{If $\eta = \frac{1}{4 \rho L}$,}
\E[f(\mxk)]  - f(\xopt) & \leq \frac{\frac{7}{8 \rho} \left[f(\x_0) - f(\xopt) \right]  + \frac{L}{2} \normsq{\x_0 - \xopt} }{\frac{1}{8 \rho} k} \\
\E[f(\mxk)]  - f(\xopt) & \leq \frac{7 \left[f(\x_0) - f(\xopt) \right]  + 4 \rho L \normsq{\x_0 - \xopt} }{k} \\
\E[f(\mxk)]  - f(\xopt) & \leq \frac{(7L/2) \normsq{\x_0 - \xopt} + 4 \rho L \normsq{\x_0 - \xopt} }{k} \\
\implies \E[f(\mxk)] - f(\xopt) & \leq \frac{4 (1+\rho) \normsq{\x_0 - \xopt}}{k} 
\end{align*}
\end{proof}
\subsection{Proof for Proposition~\ref{prop:wgc-sgc}}
\label{app:wgc-sgc}
\begin{proof}
\begin{align*}
\intertext{For the first part, we use the PL inequality which states the for all $\x$,}
2 \left[ f(\x) - f(\xopt) \right] & \leq \frac{1}{\mu} \normsq{\grad{\x}} \\
\intertext{Combining this with the WGC gives us the desired result}
\end{align*}
\begin{align*}
\intertext{For the converse, we use smoothness and the convexity of $f(\cdot)$. Specifically, for all points $a$, $b$,}
f(a) - f(b) & \geq \langle f(b), a - b  \rangle + \frac{1}{2L} \normsq{  \grad{a} - \grad{b} } & \nonumber \\
\intertext{Substituting $a = \x$ and $b = \xopt$ and rearranging,}
\normsq{ \grad{\x} } & \leq 2L \cdot \left[ f(\x) - f(\xopt)  \right]
\intertext{Combining this with the SGC gives us the desired result.}
\end{align*}
\end{proof}
\subsection{Proof for Proposition~\ref{prop:wgc-f}}
\label{app:wgc-f}
\begin{proof}
\begin{align}
\E_i \normsq{\nabla f_i(\x)} & = \frac{1}{n} \sum_{i = 1}^{n} \normsq{\nabla f_i(\x)} & \label{eq:belkin-temp} \\
\intertext{By Lipschitz continuity of $\nabla f_i(\x)$ and convexity,}
f_{i}(\x) - f_{i}(\xopt) & \geq \langle \nabla f_{i}(\xopt), \x - \xopt \rangle +  \frac{1}{2 L_i} \normsq{\nabla f_i(\x) - \nabla f_{i}(\xopt)} & \nonumber \\
\intertext{For all $i$, $\nabla f_{i}(\xopt) = \nabla f(\xopt) = 0$. Hence,}
f_{i}(\x) - f_{i}(\xopt) & \geq \frac{1}{2 L_i} \normsq{\nabla f_i(\x)} & \nonumber \\
\implies \normsq{\nabla f_i(\x)} & \leq 2 L_i \left[ f_{i}(\x) - f_{i}(\xopt) \right] & \nonumber \\
\intertext{Using Equation~\ref{eq:belkin-temp},}
\E_i \normsq{\nabla f_i(\x)} & \leq \sum_{i = 1}^{n} \left[ \frac{2 L_i}{n} \left[ f_{i}(\x) - f_{i}(\xopt) \right] \right] & \nonumber \\
& \leq \frac{2 L_{max}}{n} \sum_{i = 1}^{n} \left[ f_{i}(\x) - f_{i}(\xopt) \right] & \nonumber \\
\E_i \normsq{\nabla f_i(\x)} & \leq 2 L_{max} \left[ f(\x) - f(\xopt) \right] 
\end{align}
\end{proof}
\subsection{Proof for Lemma~\ref{lemma:sq-hinge-sgc}}
\label{app:sq-hinge-sgc}
\begin{proof}
Let $a = y \cdot x$. For the squared-hinge loss, the strong growth condition is equivalent to
\begin{align*}
\E \big[  ( 1 - \x^\top a )_+^2 \big] & \leqslant \rho \big\| \E \big[ ( 1 - \x^\top a)_+ a \big] \|^2 \\
\big\| \E \big[ ( 1 - w^\top a)_+ a \big] \big\| & \geqslant \frac{1}{\| \x_\ast\|}  \E \big[ ( 1 - \x^\top a)_+ a^\top \x_\ast \big] \\
& \geqslant \tau  \E \big[ ( 1 - \x^\top a)_+   \big] 
\end{align*}
We thus need to upper bound $\E \big[  ( 1 - w^\top a)_+^2 \big]$ by a constant $c$ times $ \big( \E \big[  ( 1 - w^\top a)_+ \big] \big)^2$. We must have $c \geqslant 1$ (as a consequence of Jensen's inequality). Then we have $\rho = c / \tau^2$. Next, we prove that if the distribution of $a$ is uniform over $\kappa$ values, then $c = \kappa$.

Consider a random variable $A \in \mathbb{R}+ $ taking $\kappa$ values $a_1,\dots,a_\kappa$ with probabilities $p_1,\dots,p_\kappa$.
Then $( \E A )^2 = \sum_{i,j} p_i p_j a_i a_j \geqslant \sum_i a_i^2 p_i^2 \geqslant \min_i p_i \sum_i a_i^2 p_i $,
\end{proof}
\subsection{Proof for Lemma~\ref{lemma:surrogate-01}}
\label{app:surrogate-01}
\begin{proof}
Let $a = y \cdot x$.
\begin{align*}
\P ( a^\top \x \leqslant 0) & \leqslant \P ( ( 1- a^\top \x)_+^2 \geqslant 1)  \\
& \leqslant \E ( 1- a^\top \x)_+^2 \\
\implies \P ( a^\top \x \leqslant 0) & \leq \E f(\x,a)
\end{align*}
\end{proof}
\section{Additional experimental results}
\label{app:exps}
In this section, we propose to use a line-search heuristic for both constant step-size SGD and its accelerated variant. For SGD, we use the line-search proposed in SAG~\cite{schmidt2017minimizing}: start with an initial estimate $\hat{L} = 1$ and in each iteration, we double the estimate when the condition $f_{k}\left( \xk - \frac{1}{\hat{L}} \nabla f_{k}(\xk) \right) \leq f_k(\xk) - \frac{1}{2 \hat{L}} \normsq{\nabla f_{k}(\xk)}$ is not satisfied. We denote this variant as SGD(LS) and the corresponding variant that uses a $1/L$ step-size as SGD(T). For the accelerated case, we use the same line-search procedure as above, but search for an appropriate value of $\rho L$. We denote the accelerated variant with and without line-search as Acc-SGD(LS) and Acc-SGD(T) respectively. 

\begin{figure*}[!ht]
\centering
        \subfigure[$\tau = 0.1$]
        {
			\includegraphics[width=0.4\textwidth]{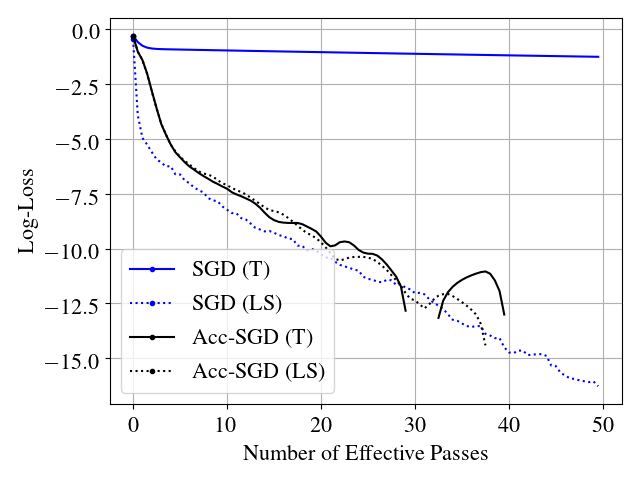}
			\label{fig:app-gamma-01}
        }        
        \subfigure[$\tau = 0.05$]
        {
			\includegraphics[width=0.4\textwidth]{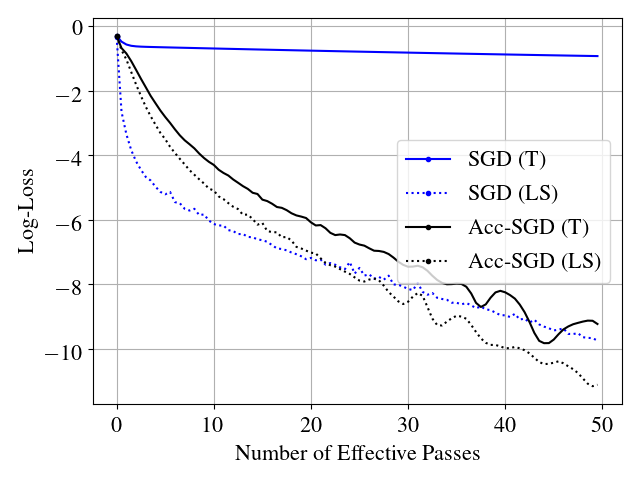}
			\label{fig:app-gamma-005}
        }      
        \\  
        \subfigure[$\tau = 0.01$]
        {
			\includegraphics[width=0.4\textwidth]{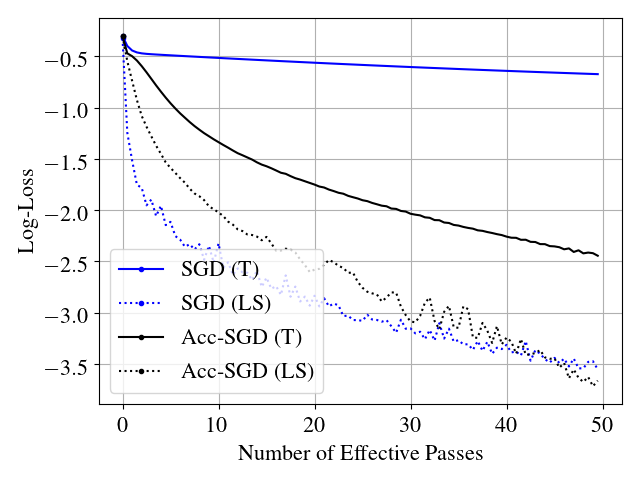}
			\label{fig:app-gamma-001}
        }        
        \subfigure[$\tau = 0.005$]
        {
			\includegraphics[width=0.4\textwidth]{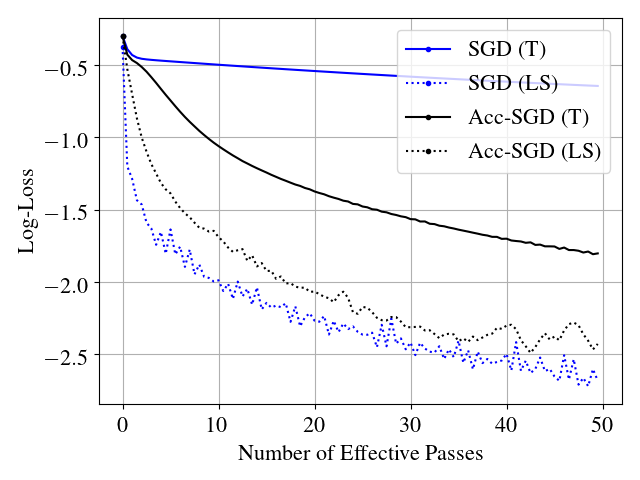}
			\label{fig:app-gamma-0005}
        }        
\caption{Comparison of SGD and variants of accelerated SGD on a synthetic linearly separable dataset with margin $\tau$. }
\label{fig:app-synthetic}
\end{figure*}    

\begin{figure*}[!ht]
\centering
        \subfigure[CovType]
        {
			\includegraphics[width=0.4\textwidth]{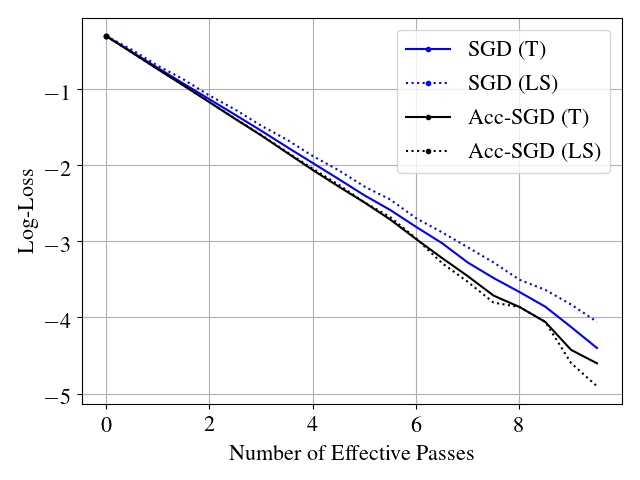}
			\label{fig:app-covtype}
        }        
        \subfigure[Protein]
        {
			\includegraphics[width=0.4\textwidth]{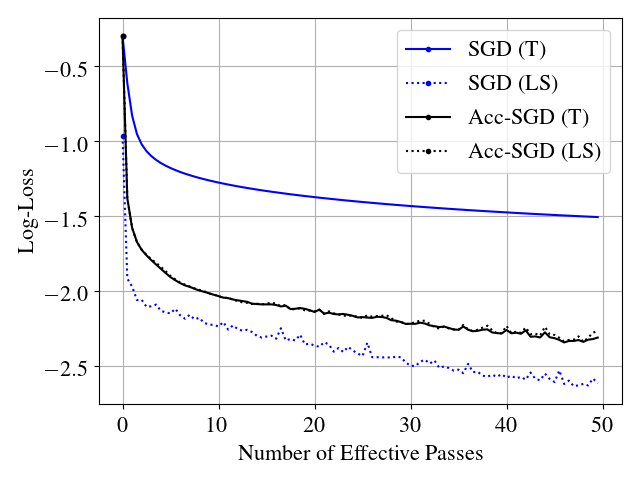}
			\label{fig:app-protein}
        }               
\caption{Comparison of SGD and accelerated SGD for learning a linear classifier with RBF features on the (a) CovType and (b) Protein datasets.}
\label{fig:app-real}
\end{figure*}

We make the following observations: (i) Accelerated SGD in conjunction with our line-search heuristic is stable across datasets.  (ii) Acc-SGD(LS) either matches or outperforms Acc-SGD(T). (iii) In some cases, SGD(LS) can result in faster empirical convergence as compared to the accelerated variants. We plan to investigate better line-search methods for both SGD~\cite{schmidt2017minimizing} and Acc-SGD~\cite{liu2009large} in the future. 

\end{document}